\documentclass[10pt,twocolumn,letterpaper]{article}

\usepackage{cvpr}              

\usepackage[table]{xcolor}

\usepackage{dsfont}
\usepackage{tcolorbox}
\usepackage{newfloat}

\definecolor{lightpink}{RGB}{251, 220, 235}
\definecolor{lightblue}{RGB}{203, 220, 235}
\definecolor{lightgreen}{RGB}{219,234,210}
\definecolor{lightred}{RGB}{255, 107, 107}
\definecolor{lightgray}{RGB}{211, 211, 211}
\definecolor{darkspringgreen}{rgb}{0.09, 0.45, 0.27}
\definecolor{ao(english)}{rgb}{0.0, 0.5, 0.0}
\definecolor{mypink}{rgb}{0.858, 0.188, 0.478}

\newcommand\blfootnote[1]{%
  \begingroup
  \renewcommand\thefootnote{}\footnote{#1}%
  \addtocounter{footnote}{-1}%
  \endgroup
}

\definecolor{ultraviolet}{HTML}{8365BA}

\newcommand{\sam}[1]{{\color{ao(english)}#1}}

\newcommand{\lab}[1]{\textit{#1}} 
\newcommand{\ourmethod}{SpeciaRL\xspace}

\newcommand{\spe}{spec.\xspace}
\newcommand{\cor}{corr.\xspace}

\newcommand{\prompt}{\mathtt{P}}
\newcommand{\netlmm}{\Phi_\mathtt{LMM}^{\theta}}
\newcommand{\netllm}{\Psi_\mathtt{LLM}}
\newcommand{\image}{I}
\newcommand{\gtlabel}{y}
\newcommand{\pred}{p}
\newcommand{\predset}{\mathcal{C}} 
\newcommand{\predcls}{c} 
\newcommand{\visualspace}{\mathcal{V}} 

\newcommand{\semanticspace}{\mathcal{S}}
\newcommand{\reward}{r}
\newcommand{\ourreward}{r^*}

\newcommand{\BoN}{\operatorname{BoN}}
\DeclareMathOperator*{\argmax}{arg\,max}
\newcommand{\wrong}{W}
\newcommand{\abstain}{A}
\newcommand{\generic}{G}
\newcommand{\lessspecific}{S^-}
\newcommand{\specific}{S}
\newcommand{\morespecific}{S^+}

\newcommand{\inlineColorbox}[2]{\begingroup\setlength{\fboxsep}{1pt}\colorbox{#1}{\hspace*{2pt}\vphantom{Ay}#2\hspace*{2pt}}\endgroup}

\newcommand{\suppmat}{\textit{Supp. Mat.}}

\usepackage{microtype}

\usepackage{multirow}
\usepackage{soul}
\usepackage{microtype}
\definecolor{cvprblue}{rgb}{0.21,0.49,0.74}
\usepackage[pagebackref,breaklinks,colorlinks,allcolors=cvprblue]{hyperref}

\def\paperID{34597} 
\def\confName{CVPR}
\def\confYear{2026}

\title{Specificity-aware reinforcement learning for fine-grained open-world classification}

\author{Samuele Angheben\textsuperscript{$1,2$} \quad Davide Berasi\textsuperscript{$1$} \quad Alessandro Conti\textsuperscript{$1$} \quad 
Elisa Ricci\textsuperscript{$1, 2$}\quad Yiming Wang\textsuperscript{$2$}\\
\small
$^1$University of Trento \quad $^2$Fondazione Bruno Kessler
}

\begin{document}


\twocolumn[{%
    \renewcommand\twocolumn[1][]{#1}%
    \vspace{-1cm}
    \maketitle
    \thispagestyle{empty}
}]

\begin{abstract}
Classifying fine-grained visual concepts under open-world settings, i.e., without a predefined label set, demands models to be both accurate and specific.
Recent reasoning Large Multimodal Models (LMMs) exhibit strong visual understanding capability but tend to produce overly generic predictions when performing fine-grained image classification.
Our preliminary analysis reveals that models do possess the intrinsic fine-grained domain knowledge. However, promoting more specific predictions (specificity) without compromising correct ones (correctness) remains a non-trivial and understudied challenge.
In this work, we investigate how to steer reasoning LMMs toward predictions that are both correct and specific.
We propose a novel specificity-aware reinforcement learning framework, SpeciaRL, to fine-tune reasoning LMMs on fine-grained image classification under the open-world setting.
SpeciaRL introduces a dynamic, verifier-based reward signal anchored to the best predictions within online rollouts, promoting specificity while respecting the model's capabilities to prevent incorrect predictions.
Our out-of-domain experiments show that SpeciaRL delivers the best trade-off between correctness and specificity across extensive fine-grained benchmarks, surpassing existing methods and advancing open-world fine-grained image classification. Code and model are publicly available at {\small \url{https://github.com/s-angheben/SpeciaRL}}.
\blfootnote{Corresponding author: {\tt sangheben@fbk.eu}.}
\end{abstract}    
\section{Introduction}
\label{sec:intro}

\begin{figure}
    \centering
    \includegraphics[trim={0 0 0 8mm}, clip, width=.99\columnwidth]{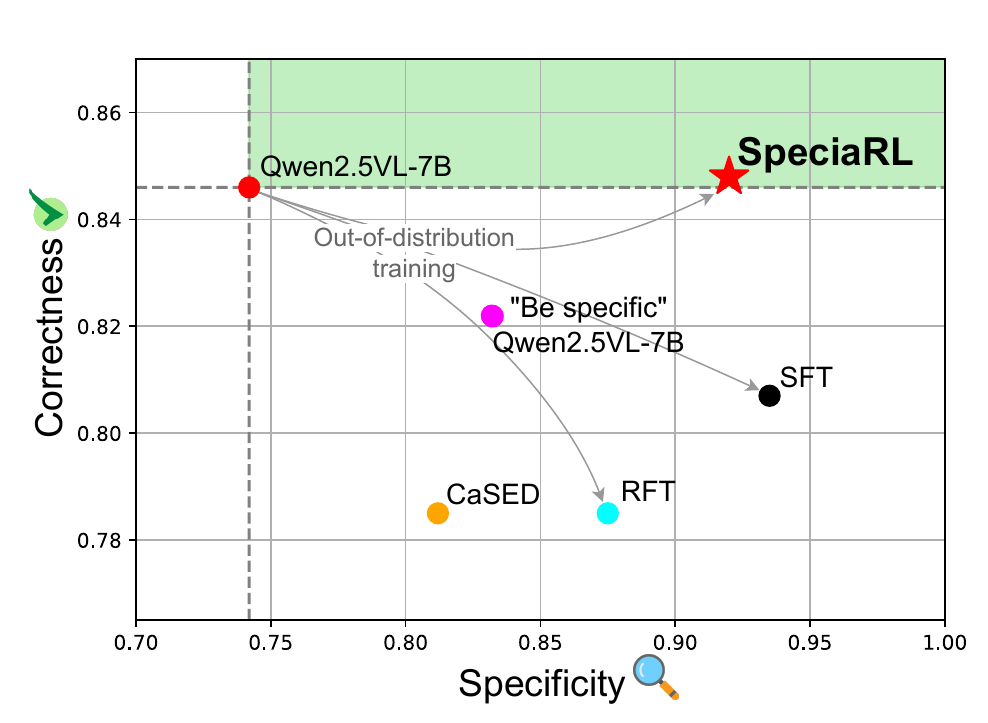}
    \vspace{-10pt}
    \caption{In open-world image classification, improving prediction specificity without compromising correctness remains challenging. Existing techniques, such as prompting to be specific, supervised fine-tuning (\textit{sft}) or reinforcement fine-tuning (\textit{rft}), promote specificity but reduce correctness. Instead, our proposed method (\textbf{\ourmethod}) significantly improves the specificity of the base Qwen2.5VL-7B model without compromising correctness. 
    Gray arrows indicate that training is performed on a single-domain (birds) dataset, which is disjoint from the domains in the test set, therefore illustrating cross-domain generalization.
    }
    \label{fig:teaser}
\end{figure}

Image classification has long been a cornerstone problem in computer vision, aiming to assign a semantic concept to the main object featured in an image~\cite{deng2009imagenet}. 
Traditional image classification models typically operate under a \textit{closed-world} setting, where all possible semantic categories are predefined within a fixed vocabulary~\cite{schmarje2021survey}.
However, in real-world environments models often need to handle emerging categories or novel concepts, highlighting the importance of \textit{open-world} classification~\cite{bendale2015towards}, which removes the fixed vocabulary assumption.
This more challenging and practically relevant setting can now be studied more effectively thanks to the emergence of large pre-trained vision–language models~\cite{radford2021clip,zhai2023siglip}.
Candidate concepts can be derived from large textual corpora~\cite{conti2023vocabulary} or directly generated by recent Large Multimodal Models (LMMs)~\cite{li2023blip2,li2024llavaNext,zhu2024minigpt4,bai2025qwen25,chen2024internvl} in response to open-ended prompts such as \textit{``What is the object in the image?''}.
Such advances have, in turn, motivated novel approaches as well as new evaluation protocols to assess the correctness of predicted concepts, addressing the unconstrained nature of LMM-generated outputs.

Recent benchmarking works~\cite{zhang2024visually, liu2024revisiting, conti2025large} extensively evaluated the classification performance of LMMs in both closed-world and open-world settings.
Focusing on the latter, Conti \textit{et al.}~\cite{conti2025large} introduced performance metrics based on large language models and textual embedding similarity, in an effort to comprehensively describe the behavior of LMMs.
The study showed that the best-performing models are recent \textit{reasoning LMMs}, such as Qwen2.5VL~\cite{bai2025qwen25}, which are trained with reasoning-enriched multimodal datasets to connect visual evidence with linguistic inference.
The study also revealed that \textit{\textbf{LMMs mostly struggle in classifying fine-grained concepts}}, with the \textbf{\textit{tendency of being overly generic}} (\eg, flower \vs daisy). 
However, na\"ively encouraging more specific predictions (\ie, high specificity) may increase the number of wrong outputs (\ie, reduced correctness).
For example, Conti \etal observed in~\cite{conti2025large} that simple prompting benefits LMMs in producing more fine-grained predictions, but at the cost of inferior correctness.
Our own experimentation also confirms this compromised correctness when promoting specificity, either by directly querying the model to ``be specific'' or by fine-tuning the model with supervised fine-tuning (\textit{sft}) or reinforcement fine-tuning (\textit{rft}), as shown in \cref{fig:teaser}.
Promoting more specific predictions requires indeed a delicate \textbf{\textit{balance between specificity and correctness}}, a non-trivial challenge that remains greatly underexplored. 

This work addresses the limitation of LMMs being overly generic on fine-grained open-world classification, aiming to improve prediction \textit{specificity} without compromising \textit{correctness}. 
Before designing the method, we conduct an in-depth inspection of the models' behavior to understand their capabilities and limitations. We analyzed the prediction distribution over several specificity levels, \eg, \textit{more specific}, \textit{specific}, \textit{less specific}, and \textit{generic}, confirming the tendency of the model being overly generic.
We further verify whether this limitation stems from a lack of domain-specific knowledge.
Interestingly, our preliminary analysis on Qwen2.5VL~\cite{bai2025qwen25}, the best-performing LMM in~\cite{conti2025large}, suggests that \textbf{\textit{the model does possess substantial prior knowledge}}, as evidenced by its strong ability to correctly identify fine-grained categories when queried multiple times, despite a few samples remaining generic or less specific.

Given these observations, we propose \ourmethod, an effective reinforcement learning method with a novel specificity-aware dynamic reward design to elicit specificity within the model's maximal capabilities. 
Intuitively, if a model’s best prediction for a given sample is inherently generic, penalizing it for lacking specificity may push it toward producing more incorrect outputs. Our \textit{sample-wise} reward is therefore \textit{dynamically} set based on the highest specificity level the LMM can achieve for that sample during multiple rollouts. This paradigm naturally blends into the GRPO~\cite{shao2024deepseekmath} algorithm without compromising computational efficiency.
\ourmethod encourages the model’s genuine reasoning capability in fine-grained visual understanding, enabling \textit{strong out-of-domain generalization} even when trained on a limited dataset from a specific domain. Empirically, \ourmethod strikes the best balance between specificity and correctness across both \textit{fine-grained} and \textit{very fine-grained} datasets, outperforming zero-shot reasoning LMMs and fine-tuned baselines.

\noindent Our main \textbf{contributions} are summarized below:
\begin{itemize}

\item We tackle the non-trivial, underexplored challenge of promoting specificity without compromising correctness in fine-grained open-world image classification. 

\item Our analysis confirms that LMMs are overly generic and provides insights on their potential and limitation.

\item We introduce \ourmethod, an online reinforcement learning method with a novel specificity-aware dynamic reward. 

\item \ourmethod achieves the best trade-off between specificity and correctness compared to existing methods.
\end{itemize}

\section{Related Work}\label{sec:related}

\noindent\textbf{Large Multimodal Models and reasoning.}
Early vision-language models primarily focused on learning a joint embedding space that aligned textual and visual representations~\cite{radford2021clip,zhai2023siglip,jia2021scaling}.
This paradigm later evolved into \emph{generative} Large Multimodal Models~\cite{alayrac2022flamingo,li2023blip2,zhu2024minigpt4,singh2022flava,wang2024qwen2,chen2024internvl, li2024llava-ov}, which connect visual features from a pretrained encoder to the input space of a Large Language Model, enabling open-ended visual question answering and visual reasoning.

Recent studies on Chain-of-Thought~(CoT) prompting~\cite{wei2022chain,kojima2022large} have demonstrated that eliciting multi-step reasoning in LMMs significantly improves their performance on several tasks. This insight has led to the development of \emph{reasoning} LMMs such as OpenAI~o1~\cite{jaech2024openai} and DeepSeek-R1~\cite{guo2025deepseek}, which are specifically fine-tuned to perform complex multi-step reasoning before providing a final answer. In this context, Reinforcement Learning has emerged as an efficient and effective post-training strategy for improving the reasoning capabilities of LMMs~\cite{abdulhai2023lmrl,ouyang2022training,stiennon2020learning,zuo2025ttrl}.

In this paper, we aim to investigate and improve the capabilities of reasoning LMMs in the specific task of open-world image classification, promoting specificity without compromising correctness.

\noindent\textbf{Evaluating LMMs as image classifiers.}
Evaluating the performance of LMMs is challenging due to their unconstrained output space.
Several comprehensive benchmarks have been introduced to test the general capabilities of LMMs~\cite{li2024seed,liu2024mmbench,li2024mvbench,zhang2025automated,fu2025mmecomprehensiveevaluationbenchmark}.  
However, the specific problem of evaluating LMMs as image classifiers, that is, assessing their ability to assign a semantic concept to a visual input, has received less attention~\cite{yue2024object, zhang2024visually,liu2024revisiting,conti2025large}.
Existing approaches reformulate classification as a multiple-choice visual question answering task~\cite{tan2025vision}, or estimate accuracy based on next-token prediction probabilities~\cite{yue2024object}.
Most relevant works include the study~\cite{snaebjarnarson2025taxonomy} on the quantification of prediction quality with hierarchical precision and recall, mapping open-ended predictions onto a predefined taxonomy through a combination of string matching and semantic similarity measures, and the benchmark~\cite{conti2025large} featuring an extensive evaluation of how various LMMs respond to the question ``What is the main object in the image?'', introducing four complementary metrics to assess different aspects of open-world prediction behavior.

Instead, we leverage the judgment of an LLM-based verifier to automatically assess and categorize the relationships between the predictions and ground-truth labels.

\noindent\textbf{Reinforcement Learning.}
Reinforcement Learning~\cite{sutton1998reinforcement} is currently the main post-training paradigm for improving the reasoning capabilities of LLMs and LMMs.
Early work on RL from Human Feedback (RLHF)~\cite{ouyang2022training,stiennon2020learning} leveraged human preference annotations as reward signals, guiding models toward being more helpful, harmless, and aligned with human preference.
More recently, RL with Verifiable Rewards (RLVR) has emerged as an effective strategy for improving reasoning~\cite{guo2025deepseek,team2025kimi,lambert2024tulu}.
Instead of relying on subjective human feedback, RLVR utilizes rule-based or programmatically verifiable reward signals obtained by directly checking model outputs against ground-truth targets. This makes RLVR particularly suitable for tasks with structured solutions, such as mathematical problem solving~\cite{shao2024deepseekmath,ying2024internlm,yang2024qwen2} and code generation~\cite{hui2024qwen2,zhang2025codedpo}. 
Notably, the Group Relative Policy Optimization (GRPO)~\cite{shao2024deepseekmath} algorithm, popularized by DeepSeek-R1~\cite{guo2025deepseek}, has shown exceptional performance. 
GRPO has been successfully applied also in vision tasks~\cite{liu2025visual,li2025think}. 
Closely related to our work, Visual-RFT~\cite{liu2025visual} applies verifiable rewards to closed-set image classification, rewarding predictions that exactly match target labels.

Given the verifiable reward assumption, RLVR has been mainly employed on tasks with structured solutions. However, recent works~\cite{su2025crossing,gunjal2025rubrics} have overcome this limitation and extended the RLVR paradigms to other domains with the help of a model-based verifier for the reward computation.

In this work, we build upon these ideas and propose a novel RL framework for open-world image classification, compatible with on-policy optimization methods such as GRPO. Our method leverages an LLM-based verifier to provide reward signals to open-ended predictions within the huge unconstrained LMMs output space.

\section{Method}\label{sec:method}

In this section, we first revisit the task of open-world image classification~\cite{conti2025large}, outlining our primary objective (Sec.~\ref{sec:method:problem_formulation}). 
Then, we introduce how to assess model predictions to quantify their specificity and correctness (\cref{sec:method:measures}).
Next, we conduct a preliminary analysis to further inspect the prediction behavior of the best-performing reasoning LMM, examining its capabilities and limitations in classifying fine-grained concepts under the open-world setting (Sec.~\ref{sec:method:preliminary}).
Finally, motivated by our preliminary findings, we introduce \ourmethod, an online RL fine-tuning approach with a novel dynamic reward design that encourages more specific predictions without increasing incorrect ones (Sec.~\ref{sec:method:method}).

\subsection{Problem formulation}
\label{sec:method:problem_formulation}

We consider the problem of classifying an image in an open-world setting, where the set of possible output classes is neither predefined nor finite.
Formally, 
we aim to learn a classifier $f: \visualspace \to \semanticspace$, which maps an image $\image \in \visualspace$ to a semantic concept $s \in \semanticspace$.
Here, $\semanticspace$ denotes a huge semantic space including all the concepts that can be expressed in natural language through concise labels.
In our setting, $f$ corresponds to a Large Multimodal Model $\netlmm$ with learnable parameters $\theta$, and $\semanticspace$ includes all the concepts that can be expressed with the model's vocabulary. 

Semantic concepts within $\semanticspace$ are not independent but semantically related following complex and hierarchical ontologies~\cite{snaebjarnarson2025taxonomy}. 
For instance, a \lab{golden-winged warbler} is a type of \lab{warbler} and more broadly a \lab{bird}. 
The unconstrained nature of the open-world setting, and the generative nature of $\netlmm$, can produce multiple possible labels that are \textit{correct at different levels of specificity}. 
As revealed in~\cite{conti2025large}, existing LMMs tend to produce \textit{\textbf{correct but generic}} predictions, particularly in fine-grained domains. 
As shown in~\cite{conti2025large}, while eliciting LMMs to be more specific through the input prompt can benefit models in producing more specific predictions, this also comes at the cost of correctness, resulting in more wrong predictions. 
How to balance the specificity and correctness remains a non-trivial challenge~\cite{snaebjarnarson2025taxonomy}.

This work focuses on addressing the limitation of LMMs being overly generic in open-world image classification. We aim to promote both \textit{specificity and correctness}, \ie, generating correct predictions with maximal specificity without harming their correctness.

\begin{figure*}[t!]
    \centering
    \includegraphics[width=\linewidth]{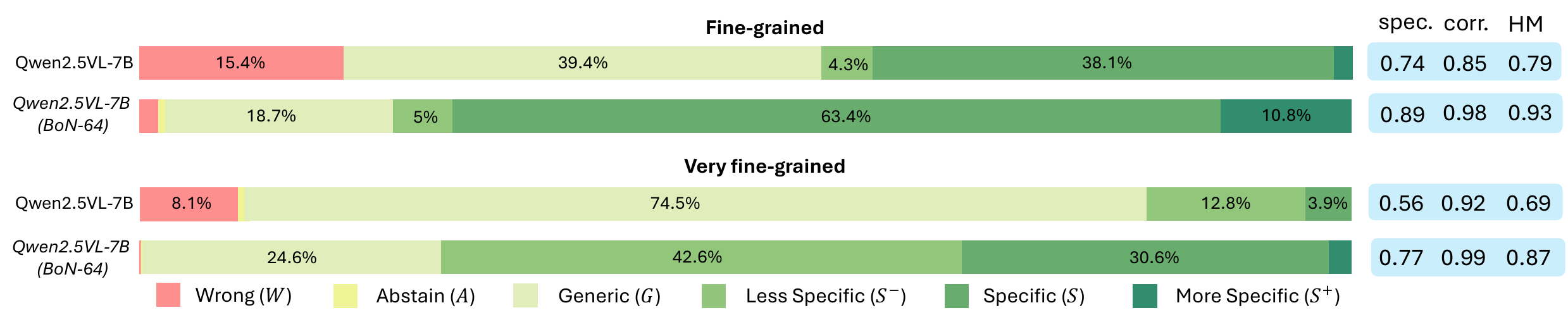}
    \vspace{-8mm}
    \caption{Predictions distribution over categories for Qwen2.5VL-7B~\cite{bai2025qwen25} and its BoN version with $N=64$ inference runs. The right side shows specificity, correctness and their harmonic mean (HM). The BoN-64 serves as an indication for the model’s potential capability.}
    \label{fig:preliminary study}
    \vspace{-5pt}
\end{figure*}

\subsection{Prediction Evaluation}\label{sec:method:measures}
Before method development, we first need to assess both the \textit{correctness} and the \textit{specificity} of model predictions.
Previous work on hierarchical classification~\cite{snaebjarnarson2025taxonomy} relies on explicit taxonomies to measure the semantic distance between labels.
However, given the open nature of our setting, we do not assume a predefined hierarchy, which is also challenging to acquire. The recent benchmark~\cite{conti2025large} introduces metrics to measure semantic similarity via both LLMs and pre-trained encoders, from which the authors broadly categorize the predictions into specific \vs generic, and correct \vs incorrect. Our extensive qualitative analysis reveals that the possible relations between a prediction $\pred$ and the fine-grained ground-truth label $\gtlabel$ are much richer: the prediction can be generic or less specific than its ground-truth label, and the model may also abstain when it judges a lack of knowledge.
However, it is not easy to manipulate these soft metrics, which are introduced to reflect the rich possibilities of model predictions.
In this work, we leverage a strong LLM as the judge categorizing the relations between the prediction $\pred$ and its fine-grained ground-truth label $\gtlabel$. 

\noindent\textbf{Prediction categorization.}
We identify a set $\predset=\{\wrong,\abstain,\generic,\lessspecific,\specific,\morespecific\}$ of six mutually exclusive categories that comprehensively cover the main possible relations:
\begin{itemize}
    \item \textbf{Wrong ($\wrong$)}: the prediction is incorrect, referring to a different concept from the target.
    
    \item \textbf{Abstain ($\abstain$)}: the prediction is a refusal to answer, which we deem as the least informative non-Wrong response a model could provide. 

    \item \textbf{Generic ($\generic$)}: the prediction is correct\sam{,} but represents a significantly broader category than the ground-truth. For example, $\pred$= dog, $\gtlabel$= samoyed.

    \item \textbf{Less Specific ($\lessspecific$)}: the prediction is correct but corresponds to a closely related parent category of the ground-truth. 
    For example, $\pred$= warbler, $\gtlabel$= golden-winged warbler.

    \item \textbf{Specific ($\specific$)}: the prediction is an exact match or a direct synonym for the ground-truth.

    \item \textbf{More Specific ($\morespecific$)}: the prediction refers to a more specific subtype or instance of the ground-truth.
    This is unlikely given that the target is a fine-grained concept, but it may occur in practice.
\end{itemize}
Note that these categories are naturally ordered from the least to the most informative as:
\begin{equation}
\wrong \prec \abstain \prec \generic \prec \lessspecific \prec \specific \prec \morespecific   
\end{equation}\label{eq:category order}
So, given two predictions $p, p'$ respectively categorized as $c, c'$, our objective considers $p$ to be better than $p'$ if $c\succ c'$.

To automatically categorize predictions, we adopt an LLM-as-a-judge approach.
We prompt a Large Language Model $\netllm$ to provide a suitable category $\predcls_y(p) \in \predset$ for a given prediction $\pred$ and ground-truth label $\gtlabel$:
\begin{equation}
    \predcls_y(p) = \netllm(<\pred,\gtlabel>, \prompt_j).
\end{equation}
Here, the judge's prompt $\prompt_j$ defines each category with precise descriptions. 
To ensure a wide range of prior knowledge and reliable evaluation of fine-grained semantics, we employ Llama3-72B~\cite{dubey2024llama3} as the judge. The exact expression of $\prompt_j$ is detailed in \suppmat

\noindent\textbf{Specificity and Correctness measures.}\label{sec:method:spec-corr definition}
We quantify the \emph{specificity} and \emph{correctness} based on the above-described categorization.
Considering a dataset $\mathcal{D}=\{(\image_i, \gtlabel_i)\}_{i=1}^n$ of $n$ labeled images, we indicate with $\predcls_i$ the category of prediction $\netlmm(\image_i,\prompt_c)$ and with $n_\wrong=\#\{i \, | \,c_i=\wrong\}_{i=1}^n$ the number of Wrong ($\wrong$) predictions.
We define \textit{correctness} as the percentage of non-Wrong predictions:
\begin{equation}\label{eq:def correctness}
    \text{correctness} = 1 - \frac{n_\wrong}{n}.
\end{equation}
To measure the specificity, we assign a specificity score $s(c)$ to each non-Wrong category as follows:
\begin{equation}\label{eq:specificity score}
    s(\abstain)=1, s(\generic)=2, s(\lessspecific)=3, s(\specific)=s(\morespecific)=4.
\end{equation}
Intuitively, consider the path over the categories from the root $\abstain$ to the leaf $\morespecific$.
The score in \cref{eq:specificity score} is the length of the intersection between the path from the root to the prediction's category $c$ and the path from the root to the ground-truth category $\specific$.
In other words, this can be seen as the amount of information provided by the prediction about the ground-truth concept.
We then define \textit{specificity} as the average normalized score over the non-Wrong predictions:
\begin{equation}\label{eq:def specificity}
\text{specificity} = \frac{1}{n-n_\wrong}\sum_{c_i \neq \wrong}\frac{s(\predcls_i)}{4}.
\end{equation}
Note that specificity lies in $[0, 1]$ and it is $0.5$ if all the correct predictions are Generic.
Finally, we consider the harmonic mean (HM) as a quantitative measure of overall performance:
\begin{equation}
\mathrm{HM} = 2\frac{\text{specificity} \times \text{correctness}}{\text{specificity} + \text{correctness}}.
\end{equation}

\subsection{On LMMs being overly generic}\label{sec:method:preliminary}

We conduct a preliminary study to gain a detailed understanding of the models' prediction behaviors in terms of correctness and specificity, aiming to identify their capabilities and limitations. We base our analysis on recent reasoning LMMs as they demonstrate the best performance on open-world image classification~\cite{conti2025large}. 

\begin{figure*}[t!]
    \centering
    \includegraphics[width=\linewidth]{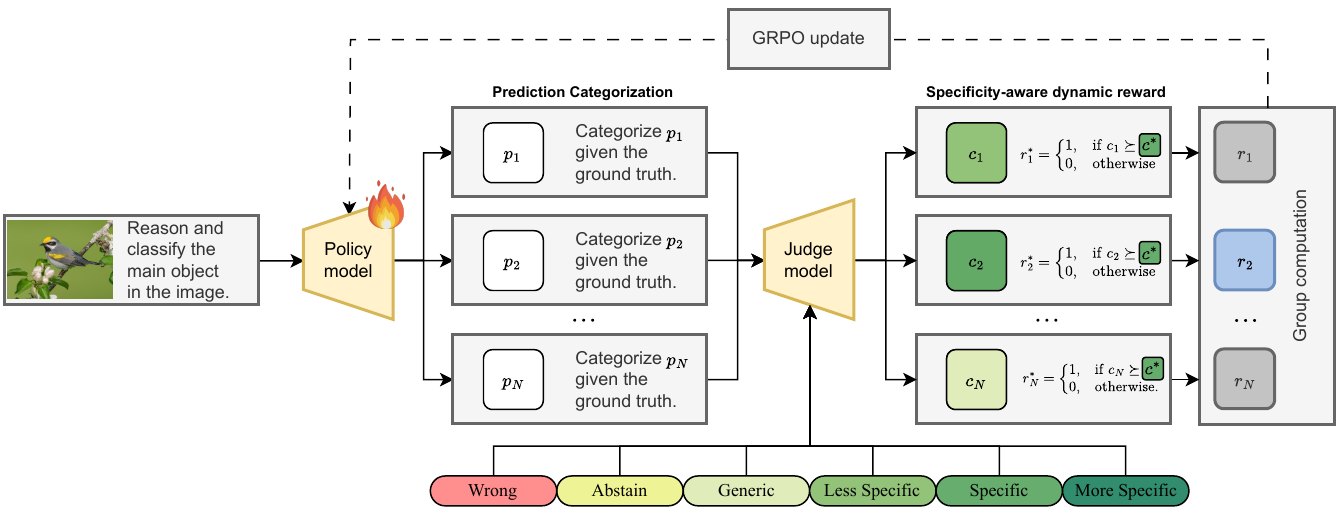}
    \vspace{-20pt}
    \caption{Overview of \textbf{\ourmethod}
    Given an input image $I$, the policy model generates $N$ open-ended predictions $\{\pred_1, \dots, \pred_N\}$. Each prediction is categorized by a judge model (LLM verifier) as wrong or correct at different levels of specificity with respect to the ground-truth. A verifiable reward $r_i^*$ is then assigned according to whether the prediction's category $c_i$ meets the adaptive reference level $\predcls^*$, which is defined based on the best prediction within the $N$ rollouts. The resulting graded rewards are aggregated through a Group Relative Policy Optimization (GRPO) update to reinforce policies that are maximally specific while remaining correct.}
    \label{fig:main}
    \vspace{-5pt}
\end{figure*}
\noindent\textbf{Experiment setting.} 
We use the \textit{\textbf{fine-grained}} set from \cite{conti2025large}, consisting of fine-grained image classification benchmarks where classes belong to a shared superclass and/or are challenging to distinguish. This includes Flowers102~\cite{nilsback2008flowers} (flowers), Food101~\cite{bossard2014food} (food), and OxfordPets~\cite{parkhi2012pets} (animals).
We also consider the \textit{\textbf{very fine-grained}} set, where categories are not only within the same subclass but also highly difficult to differentiate. This includes StanfordCars~\cite{krause20133cars}, where labels specify car brands, models, and years of production, and FGVCAircraft~\cite{maji2013aircraft}, which categorizes aircraft models.

Each image $\image$ in a dataset is associated with a human-annotated ground-truth label $\gtlabel\in \semanticspace$. The model's prediction $\pred\in\semanticspace$ is obtained by prompting the reasoning LMM $\netlmm$: 
\begin{equation}
    p=\netlmm(\image,\prompt_c)
\end{equation}
where $\prompt_c$ is a text prompt querying the model to classify the main object in the images by first reasoning on the input and then including the final prediction in \texttt{<answer>} tags. We report the exact expression of our prompt in the \suppmat. 
Specifically, we consider Qwen2.5VL-7B~\cite{bai2025qwen25} as $\netlmm$, which can perform visual understanding with linguistic reasoning and follow the thought-answer template for the output.

\noindent\textbf{How specific are model predictions?}
\Cref{fig:preliminary study} (Row I\&III) shows the percentage of predictions within each category, as well as their correctness and specificity scores. The model predictions are mostly correct, but with a clear tendency towards being generic, as already observed in~\cite{conti2025large}. This inclination is more evident in the case of the very fine-grained set (Row III), where almost 75\% of the predictions are Generic.

\noindent\textbf{Does the model have prior domain knowledge?}
We wonder whether the tendency to be generic is due to the lack of domain-specific knowledge. We evaluate this aspect by considering the best prediction over $N$ rollouts based on the intuition that the model may possess the prior knowledge to produce 
better predictions, but it may be inefficient in sampling the correct reasoning path in a single attempt~\cite{yue2025does}.
Specifically, we define the Best-of-N~(BoN) prediction for a given sample $(I, y)$ as the prediction within $N$ generations $\{p_1, \dots, p_N\}$ with the most informative category:
\begin{equation}
    \BoN_y(I) = \argmax_{\pred \in\{\pred_1, \dots, \pred_N\}} \netllm( <\pred,\gtlabel>, \prompt_j).
\label{eq:BoN definition}
\end{equation}
We set $N=64$, a computationally reasonable value that is sufficiently large to provide a reliable upper bound on the model classification capability.

As shown in \cref{fig:preliminary study} (Row II\&IV), the best prediction within 64 rollouts (BoN-64) shows significantly \textit{greater specificity and correctness} compared to one-time inference, as evident in both the distribution over categories and the metric scores. This suggests that the model \textit{does possess the prior knowledge} to be substantially more precise, despite its generic tendency. We hypothesize that this might be due to the bias inherited from the pretraining distribution, where generic concepts are much more frequent than specific ones. 
On the other hand, the BoN-64 results reveal that, even at its best, the LMM still produces a decent portion of Generic or Less Specific predictions, particularly in very fine-grained cases. This suggests some samples still lie outside the model's capabilities.
These findings raise a compelling question: Is it possible to steer the model towards more specific predictions, approaching the BoN-64 performance, without pushing it over its actual potential, to avoid increasing incorrect answers?

\subsection{Specificity-aware Reinforcement Learning}\label{sec:method:method}
In our preliminary analysis, we observed that pretrained LMMs lack specificity in their classification predictions, tending toward generic responses.
Importantly, we noted this is not due to a lack of prior knowledge. 
For this reason, we propose a fine-tuning strategy that guides the model's behavior to optimize its capability to provide correct and specific predictions.
Given that the base model possesses a good level of prior knowledge, 
we do not aim to inject new knowledge into the model, but rather, we seek to improve its sampling efficiency and reasoning capabilities.
For this reason, we adopt a reinforcement learning approach, which is highly effective in steering the model's behavior and increasing reasoning performance~\cite{guo2025deepseek,team2025kimi,lambert2024tulu}. 

\noindent\textbf{Reinforcement Learning with Verifiable Rewards}
enables fine-tuning a model using a simple rule-based reward signal on tasks where a prediction is directly verifiable against the correct answer.
Originally proposed to improve LLMs' performance on language tasks such as math and coding~\cite{guo2025deepseek}, 
RLVR has recently been shown to be effective in vision tasks as well~\cite{liu2025visual,li2025think}.
Among RLVR algorithms, we adopt GRPO~\cite{shao2024deepseekmath} for its efficiency and effectiveness.
At its core, GRPO generates groups of diverse outputs $\{p_1, \dots, p_N\}$ and optimizes to incentivize responses with higher rewards within each group.
The core of RLVR is the definition of the reward signal.
Given label $\gtlabel$ and a model prediction $\pred = \netlmm(\image, \prompt_c)$, a standard verifiable reward is defined as:
\begin{equation}
\reward(\pred, \gtlabel) =
\begin{cases}
1, & \text{if } \pred = \gtlabel, \\
0, & \text{otherwise.}
\end{cases}
\label{eq:static reward}
\end{equation}
Note that this simple definition assumes the possibility of directly comparing a prediction against the target solution.

\noindent\textbf{Specificity-aware dynamic reward.}
Considering that, in an open-world setting, a prediction can be correct at different specificity levels, the standard reward could risk pushing the model to be \textit{overly specific at the cost of correctness}.
We therefore design a custom reward signal suited for open-world classification.
For a given sample $(\image, \gtlabel)$, we argue that any correct prediction, even if it does not match the ground-truth label, should be positively rewarded if it achieves the model's maximum potential.
Formally, we use the best prediction category within $N$ runs $\predcls_{best}\!=\!\predcls_y(\BoN_y(I))$ to define a minimal specificity requirement $c^*\in\mathcal{C}$ to be positively rewarded, accounting for the corner cases $c_{best}\in\{\morespecific,\wrong\}$:
\begin{equation}
\predcls^* =
\begin{cases}
S, & \text{if }  \predcls_{best} = S^+ \\
A, & \text{if } \predcls_{best} = W \\
\predcls_{best}, & \text{otherwise}. 
\end{cases}
\label{eq:def c^*} 
\end{equation}
Our sample-specific reward for a prediction $\pred$ is defined as:
\begin{equation}
\ourreward_I(\pred, \gtlabel) =
\begin{cases}
1, & \text{if } \predcls_y(p) \succeq \predcls^* \\
0, & \text{otherwise.}
\end{cases}
\label{eq:our reward} 
\end{equation}
This reward is therefore positive when the prediction is Specific, More Specific, or at least as informative as the best prediction within the current model's capability.
For example, a Generic prediction receives a positive reward if the BoN prediction is also Generic, but it is not rewarded if the BoN is Specific or Less Specific. Wrong predictions always receive reward 0.

We compute the BoN prediction in an \emph{online} manner, that is, with the current weights of the model. Specifically, we use the $N$ rollouts of the GRPO algorithm. This makes the reward computation efficient, as it does not require any additional generations compared to the static reward in \cref{eq:static reward}.

\begin{table*}[t!]
\caption{Open-world image classification results of \inlineColorbox{lightblue}{zero-shot} methods and models \inlineColorbox{lightgreen}{fine-tuned out-of-domain}.
We report the ratio of the predictions within categories assigned by the LLM verifier, our measures of specificity and correctness and the harmonic mean of these two (HM). Results are averaged over all datasets within the \textbf{fine-grained} and \textbf{very fine-grained} sets. For reference, we report the performance of \inlineColorbox{gray!15}{inference out of 64 runs (BoN-64)}. Best in \textbf{bold}; second best \underline{underlined}.}
\centering
\small
\setlength{\tabcolsep}{3pt}
\resizebox{\textwidth}{!}{%
\begin{tabular}{lcccccc|ccc||cccccc|ccc}
\toprule
& \multicolumn{9}{c||}{\cellcolor{gray!15}\textbf{Fine-grained}} & \multicolumn{9}{c}{\cellcolor{gray!15}\textbf{Very fine-grained}} \\
\cmidrule(lr){2-10} \cmidrule(lr){11-19}
& \multicolumn{6}{c|}{Prediction categorization} & \multicolumn{3}{c||}{Metrics} & \multicolumn{6}{c|}{Prediction categorization} & \multicolumn{3}{c}{Metrics} \\
\cmidrule(lr){2-7} \cmidrule(lr){8-10} \cmidrule(lr){11-16} \cmidrule(lr){17-19}
Model & $\morespecific$ & $\specific$ & $\lessspecific$ & $\generic$ & $\abstain$ & $\wrong$ & \spe~$\uparrow$ & \cor~$\uparrow$ & HM~$\uparrow$ & $\morespecific$ & $\specific$ & $\lessspecific$ & $\generic$ & $\abstain$ & $\wrong$ & \spe~$\uparrow$ & \cor~$\uparrow$ & HM~$\uparrow$ \\
\midrule
\cellcolor{lightblue}CaSED~\cite{conti2023vocabulary} & 0.0\% & 43.7\% & 10.6\% & 24.2\% & 0.0\% & 21.5\% & 0.812 & 0.785 & 0.797 & 0.0\% & 0.9\% & 13.8\% & 56.0\% & 0.0\% & 29.3\% & 0.56 & 0.707 & 0.612 \\
\cellcolor{lightblue}InternVL2.5-4B~\cite{chen2024expanding} & 0.0\% & 11.4\% & 1.5\% & 54.4\% & 8.4\% & 24.1\% & 0.554 & 0.759 & 0.639 & 0.0\% & 0.1\% & 1.2\% & 62.7\% & 5.5\% & 30.5\% & 0.486 & 0.695 & 0.571\\
\cellcolor{lightblue}InternVL2.5-8B~\cite{chen2024expanding} & 0.7\% & 16.7\% & 3.3\% & 30.6\% & 20.7\% & 27.9\% & 0.575 & 0.721 & 0.624 & 0.0\% & 1.2\% & 5.7\% & 54.5\% & 15.6\% & 22.9\% & 0.476 & 0.771 & 0.589\\
\cellcolor{lightblue}Qwen2.5VL-3B~\cite{bai2025qwen25} & 0.8\% & 17.3\% & 2.7\% & 53.4\% & 4.2\% & 21.5\% & 0.608 & 0.785 & 0.685 & 0.1\% & 1.1\% & 3.9\% & 75.1\% & 2.4\% & 17.4\% & 0.511 & 0.826 & 0.631 \\
\cellcolor{lightblue}Qwen2.5VL-7B~\cite{bai2025qwen25} & 1.4\% & 38.1\% & 4.3\% & 39.4\% & 1.4\% & 15.4\% & 0.742 & \underline{0.846} & 0.790 & 0.1\% & 3.9\% & 12.8\% & 74.5\% & 0.6\% & 8.1\% & 0.555 & \textbf{0.919} & 0.692 \\
\cellcolor{lightblue}Qwen2.5VL-7B (``Be specific'') & 2.1\% & 49.1\% & 6.2\% & 22.4\% & 3.4\% & 16.8\% & 0.816 & 0.832 & 0.822 & 0.3\% & 12.5\% & 29.3\% & 45.6\% & 1.3\% & 11.0\% & 0.652 & \underline{0.89} & 0.751 \\
\midrule
\midrule
\cellcolor{lightgreen}Qwen2.5VL-7B (\textit{sft}) & 2.4\% & 64.4\% & 7.6\% & 6.0\% & 0.3\% & 19.3\% & \textbf{0.935} & 0.807 & \underline{0.866} & 0.5\% & 22.5\% & 50.8\% & 11.8\% & 0.1\% & 14.3\% & 0.789 & 0.857 & 0.814 \\
\cellcolor{lightgreen}Qwen2.5VL-7B (\textit{rft}) & 4.6\% & 52.2\% & 5.0\% & 16.2\% & 0.0\% & 21.5\% & 0.875 & 0.785 & 0.825 & 1.2\% & 24.7\% & 53.9\% & 3.5\% & 0.0\% & 16.7\% & \textbf{0.825} & 0.833 & \underline{0.821} \\
\midrule
\cellcolor{lightgreen}\textbf{\ourmethod-7B} & 5.6\% & 63.4\% & 5.1\% & 10.7\% & 0.0\% & 15.2\% & \underline{0.920} & \textbf{0.848} & \textbf{0.883} & 1.0\% & 25.2\% & 54.2\% & 5.1\% & 0.0\% & 14.5\% & \underline{0.818} & 0.855 & \textbf{0.830} \\
\midrule
\rowcolor{gray!10}\textit{Qwen2.5VL-7B (BoN-64)} & \textit{10.8\%} & \textit{63.4\%} & \textit{5.0\%} & \textit{18.7\%} & \textit{0.6\%} & \textit{1.6\%} & \textit{0.889} & \textit{0.984} & \textit{0.933} & \textit{1.9\%} & \textit{30.6\%} & \textit{42.6\%} & \textit{24.6\%} & \textit{0.1\%} & \textit{0.2\%} & \textit{0.77} & \textit{0.998} & \textit{0.868} \\
\bottomrule
\end{tabular}
}
\vspace{-5pt}

\label{tab:complete_mean_comparison}
\end{table*}

\section{Experiments}\label{sec:exp}
In this section, we first describe the experimental setup, specifying the datasets, evaluation protocols and training details. Then, we present comparative analysis against state-of-the-art methods, supported by qualitative examples, proving \ourmethod achieves the best specificity-correctness trade-off (\cref{sec:exp:comparison}).
Finally, we show ablation studies on our key design choices on the dynamic reward (\cref{sec:exp:ablation}).

\noindent\textbf{Datasets.} 
For the evaluation, we use the same \textit{\textbf{fine-grained}} and \textit{\textbf{very fine-grained}} datasets as detailed in Sec.~\ref{sec:method:preliminary}.
For training, we randomly select 3000 samples from the CUB dataset~\cite{wah2011caltech}, a bird species classification dataset with \textit{\textbf{fine-grained}} annotations.
Note that training and testing data are from different domains.
All evaluations are therefore conducted in an \textit{out-of-domain} setting to assess generalization and reasoning capabilities rather than memorization.

\noindent\textbf{Evaluation metrics.}
Model predictions are obtained using the same prompting strategy described in \cref{sec:method:preliminary}. We evaluate both \textit{specificity} and \textit{correctness}, as well as their harmonic mean (HM) defined in \cref{sec:method:spec-corr definition}. The HM captures how well a model balances specificity and correctness, providing a single scalar measure of overall performance.
For completeness, we also report the proportion of predictions assigned to each category by the LLM judge.
To position our \ourmethod in the literature, we also follow the general-purpose evaluation protocol introduced in~\cite{conti2025large}, assessing performance using LLM evaluation, string matching and semantic similarity between model's outputs and ground-truth labels. While useful for indicating overall performance, these metrics are not specifically designed to quantify specificity and correctness.

\noindent\textbf{Training details.}
We use Qwen2.5VL-7B as the base model.
Training is performed with Group Relative Policy Optimization with the following configuration: number of rollouts per sample: $N = 10$, training batch size: $256$, learning rate: $\eta = 3 \times 10^{-5}$, total training epochs: $15$, and KL penalty coefficient: $\lambda = 0.01$.
For reward computation, we use Qwen3-30B-A3B-Instruct-2507-FP8~\cite{qwen3technicalreport} as the external LLM judge. Note that this is different from the Llama3-72B~\cite{dubey2024llama3} model used for evaluation. This distinction avoids the influence of family-specific biases and ensures a fair evaluation. 
All reinforcement learning experiments are implemented using the Verl framework~\cite{sheng2024hybridflow}.

\subsection{Main comparison}
\label{sec:exp:comparison}
\noindent\textbf{Baselines.}
We compare \ourmethod against both zero-shot and training-based baselines.
For zero-shot methods, we consider both the retrieval-based CaSED~\cite{conti2023vocabulary}, which exploits CLIP~\cite{radford2021clip} to retrieve candidate concepts from web-scale textual corpus, and state-of-the-art reasoning LMMs, including Qwen2.5VL-(3B \& 7B)~\cite{bai2025qwen25} and InternVL2.5-(4B \& 8B)~\cite{chen2024internvl}, to examine performance across architectures and scales. 
We also elicit Qwen2.5VL-7B to be specific in its predictions via prompting (\textit{``Be specific''}).

For training-based methods, we consider fine-tuning the strongest reasoning LMM Qwen2.5VL-7B, with supervised fine-tuning (\textit{sft}) and reinforcement fine-tuning (\textit{rft}).
Specifically, Qwen2.5VL-7B (\textit{sft}) performs supervised fine-tuning by cross-entropy loss on a custom dataset of high-quality reasoning traces constructed similarly to \cite{zelikman2024star,zhang2025improve}.
Precisely, for each training sample, we prompt the base model to generate a reasoning–answer pair leading to the correct ground-truth label.
We opt to use the same base model to avoid introducing extra knowledge into the model.
Qwen2.5VL-7B(\textit{rft}) is trained with GRPO using the common static reward signal, which assigns positive feedback only to predictions matching the ground-truth. In our setting, this corresponds to a reward $1$ when the prediction is categorized as Specific ($S$) or More Specific ($S^+$), $0$ otherwise.

Finally, we report the Best-of-64 performance defined in \cref{sec:method:preliminary} as an empirical upper bound on the base model’s potential capabilities.

\noindent\textbf{Quantitative results.}
As shown in \cref{tab:complete_mean_comparison}, the retrieval-based method CaSED~\cite{conti2023vocabulary} achieves promising specificity, while all zero-shot reasoning LMMs are limited in specificity as they produce mostly Generic predictions.
Eliciting specificity through the prompt, \ie Qwen2.5VL-7B(\textit{``Be specific''}), reduces Generic predictions, but also leads to more Wrong predictions.
On the other hand, all training-based approaches substantially improve specificity. 
Yet, on balancing specificity and correctness, \ourmethod achieves the best performance with the highest HM across both test groups, with less compromise on correctness.
Notably, on the \textbf{\textit{fine-grained}} dataset, \ourmethod improves both specificity and correctness compared to the base Qwen2.5VL-7B model.
Please refer to \suppmat \ for the performance on each individual dataset and for the in-domain evaluation on the CUB~\cite{wah2011caltech} test split, where the RL-based variants achieve the best overall trade-off between specificity and correctness, surpassing BoN-64 in harmonic mean.

\begin{table}[t!]
\caption{Comparison against state-of-the-art methods following the evaluation protocol of~\cite{conti2025large}. Key: TI:Text Inclusion; LI: Language Inclusion; SS: Semantic Similarity; CS: Concept Similarity. Best in \textbf{bold}; second best \underline{underlined}.
}
\centering
\resizebox{\columnwidth}{!}{%
\begin{tabular}{lcccc||cccc}
\toprule
& \multicolumn{4}{c||}{\cellcolor{gray!15}\textbf{Fine-grained}} & \multicolumn{4}{c}{\cellcolor{gray!15}\textbf{Very fine-grained}} \\
\cmidrule(lr){2-5} \cmidrule(lr){6-9}
\textbf{Model} & \textbf{TI}$\uparrow$ & \textbf{LI}$\uparrow$ & \textbf{SS}$\uparrow$ & \textbf{CS}$\uparrow$ & \textbf{TI}$\uparrow$ & \textbf{LI}$\uparrow$ & \textbf{SS}$\uparrow$ & \textbf{CS}$\uparrow$ \\
\midrule
\rowcolor{lightblue}\multicolumn{9}{l}{\textit{Retrieval-based baselines}} \\
CASED & 27.4 & 46.6 & 60.7 & 61.7 & 0.7 & 47.1 & 38.5 & 38.5 \\
CLIP retrieval & 32.4 & 45.4 & 42.9 & 65.4 & 7.0 & 18.1 & 39.7 & 56.1 \\
\midrule
\rowcolor{lightblue}\multicolumn{9}{l}{\textit{Non-reasoning LMMs}} \\
IDEFICS2 8B & 3.0 & 49.9 & 38.0 & 41.7 & 0.0 & 67.0 & 29.6 & 33.6 \\
INSTRUCTBLIP Vic 7B & 10.4 & 48.8 & 35.6 & 47.2 & 0.0 & 61.0 & 30.0 & 34.3 \\
INTERNVL2 2B & 14.9 & 47.0 & 31.6 & 50.7 & 0.7 & 32.9 & 33.1 & 43.9 \\
INTERNVL2 4B & 16.2 & 44.4 & 32.0 & 52.0 & 1.7 & 36.8 & 33.8 & 44.2 \\
INTERNVL2 8B & 22.3 & 46.7 & 34.8 & 56.7 & 2.3 & 32.5 & 36.0 & 49.4 \\
LLAVA-1.5 7B & 8.4 & 46.5 & 28.2 & 44.8 & 0.0 & 41.0 & 28.6 & 37.6 \\
LLAVA-NEXT Mist 7B & 26.8 & 43.7 & 35.3 & 60.1 & 1.4 & 47.2 & 34.2 & 46.9 \\
LLAVA-NEXT Vic 7B & 16.9 & 44.5 & 32.2 & 53.2 & 1.3 & 42.2 & 34.5 & 46.1 \\
LLAVA-OV Qwen2 0.5B & 6.0 & 42.7 & 38.5 & 43.3 & 0.6 & 65.6 & 30.5 & 37.1 \\
LLAVA-OV Qwen2 7B & 6.4 & 40.4 & 39.0 & 43.8 & 0.0 & 76.7 & 31.9 & 32.4 \\
PHI-3-VISION & 13.4 & 49.1 & 31.8 & 47.2 & 0.2 & 45.0 & 28.9 & 36.0 \\
QWEN2VL 2B & 35.7 & 62.5 & 40.7 & 63.4 & 12.9 & 60.7 & 45.1 & 62.3 \\
QWEN2VL 7B & 34.6 & 64.0 & 39.2 & 62.9 & 0.8 & 63.0 & 34.5 & 43.4 \\
\midrule
\rowcolor{lightblue}\multicolumn{9}{l}{\textit{Reasoning LMMs}} \\
INTERNVL2.5 2B & 12.2 & 38.6 & 27.5 & 47.0 & 0.8 & 52.4 & 31.6 & 41.5 \\
INTERNVL2.5 4B & 17.2 & 48.2 & 32.8 & 52.3 & 0.5 & 55.6 & 31.4 & 39.7 \\
INTERNVL2.5 8B & 17.9 & 50.9 & 32.8 & 53.5 & 1.6 & 59.9 & 32.1 & 40.4 \\
QWEN2.5VL 3B & 44.3 & 63.9 & 41.6 & 69.3 & 9.4 & 58.9 & 39.9 & 58.5 \\
QWEN2.5VL 7B & 58.7 & 74.2 & 47.0 & 78.9 & 16.4 & 70.4 & 45.8 & 68.4 \\
\midrule
\rowcolor{lightgreen}\multicolumn{9}{l}{\textit{Reasoning LMMs - fine-tuned out-of-distribution}} \\
Qwen2.5VL-7B (\textit{sft}) & 60.0 & 73.8 & 47.8 & 80.1 & 17.1 & \textbf{71.1} & 47.3 & 71.1 \\
Qwen2.5VL-7B (\textit{rft}) & \underline{62.0} & \textbf{74.8} & \underline{48.4} & \underline{80.6} & \underline{21.9} & \underline{68.2} & \underline{49.5} & \underline{74.0} \\
\midrule
\textbf{\ourmethod-7B} & \textbf{62.7} & \underline{74.4} & \textbf{49.2} & \textbf{81.1} & \textbf{24.9} & 63.8 & \textbf{50.5} & \textbf{75.4} \\
\bottomrule
\end{tabular}
}
\vspace{-5pt}
\label{tab:lmm_category_results_split}
\end{table}

Moreover, \cref{tab:lmm_category_results_split} reports the performance of \ourmethod following the evaluation protocol of a recent LMM benchmark on open-world image classification~\cite{conti2025large}. On this general-purpose benchmark, \ourmethod achieves state-of-the-art performance on three out of four metrics on both the \textit{\textbf{fine-grained}} and the \textbf{\textit{very fine-grained}} test groups, 
further validating its advantage against existing methods.



\begin{figure}
    \centering
    \includegraphics[width=0.99\columnwidth]{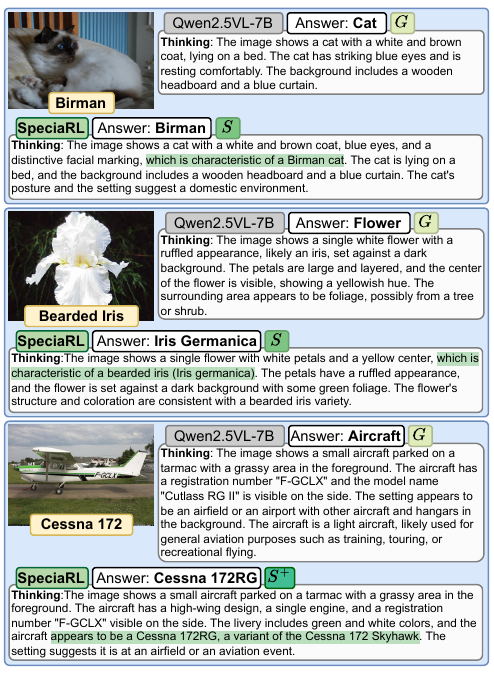}
    \vspace{-3mm}
    \caption{Qualitative examples of the think-answer output from the base model Qwen2.5VL-7B and our \ourmethod, which steers the reasoning traces towards more specific prediction.}
    \label{fig:qualitatives}
    \vspace{-5pt}
\end{figure}

\noindent\textbf{Qualitative results.}
\Cref{fig:qualitatives} presents the model outputs from the base model and our \ourmethod. For each sample, we visualize both the generated answer and the associated reasoning trace, together with the prediction category evaluated by the judge LLM. 
Consistent with the quantitative results, \ourmethod generally produces more specific and fine-grained predictions than the base model.
While both models are able to capture fine visual details in their thinking process, only \ourmethod uses these details to deduce a fine-grained class, as highlighted in green in the reasoning traces.
This suggests that our reinforcement learning strategy not only encourages specificity in the final prediction but also enhances the quality and goal-orientation of the reasoning process itself.

\subsection{Ablation studies}\label{sec:exp:ablation}
To justify the key design choices of our reward, we ablate the impact of the \textit{specificity-aware dynamic reward} and the \textit{number $N$ of online rollouts}. All ablations are conducted with Qwen2.5VL-7B and evaluated on the \textbf{\textit{fine-grained}} set. Additional studies in \suppmat \ cover training-data configurations (domain, scale, and mixed-domain setups), different on-policy RL algorithms, judge robustness and sensitivity.

\noindent\textbf{Different verifiable rewards settings.} 
\begin{table}[t!]
\caption{\ourmethod against \emph{rft} with different static reward rules. Best in \textbf{bold}.} 
\vspace{-6pt}
\centering
\small
\setlength{\tabcolsep}{3pt}
\resizebox{\columnwidth}{!}{
\begin{tabular}{lcccccc|ccc}
\toprule
& \multicolumn{6}{c|}{Prediction categorization} & \multicolumn{3}{c}{Metrics}\\
\cmidrule(lr){2-7} \cmidrule(lr){8-10}
Model & $\morespecific$ & $\specific$ & $\lessspecific$ & $\generic$ & $\abstain$ & $\wrong$ & \spe~$\uparrow$ & \cor~$\uparrow$ & HM~$\uparrow$  \\
\midrule
$S^{+}\&S$(1) & 4.6\% & 52.2\% & 5.0\% & 16.2\% & 0.0\% & 21.5\% & 0.875 & 0.785 & 0.825 \\
$S^{+}\&S$(1)$S^{-}$(0.75) & 4.9\% & 62.6\% & 5.6\% & 10.4\% & 0.0\% & 16.4\% & 0.919 & 0.836 & 0.875 \\
$S^{+}\&S$(1)$S^{-}$(0.75)$G$(0.5) & 3.6\% & 61.1\% & 5.1\% & 17.7\% & 0.0\% & 12.6\% & 0.884 & \textbf{0.874} & 0.878 \\
$S^{+}\&S$(1)$S^{-}$(0.75)$G$(0.5)$A$(0.25) & 1.4\% & 63.9\% & 6.7\% & 11.5\% & 0.0\% & 16.5\% & 0.911 & 0.835 & 0.871 \\
\ourmethod-7B~(dynamic reward) & 5.6\% & 63.4\% & 5.1\% & 10.7\% & 0.0\% & 15.2\% & \textbf{0.920} & 0.848 & \textbf{0.883} \\
\bottomrule
\end{tabular}
}
\label{tab:ablation_reward}
\end{table}
We compare our specificity-aware dynamic reward against four different static rewards.
Starting from the \emph{rft} baseline ``$S^{+}\&S$(1)'' giving reward 1 to $\specific$ and $\morespecific$ predictions, we give credit to less informative categories with a positive reward matching the specificity score as defined in \cref{eq:specificity score}. 
As shown in \cref{tab:ablation_reward}, 
\ourmethod achieves the best harmonic mean among all the static-reward variants. Interestingly, the standard binary reward ``$S^{+}\&S$(1)'' performs worst compared to the other alternatives, highlighting the importance of rewarding correct predictions that are less informative than the ground-truth.

\noindent\textbf{Impact of best of $N$ rollouts.}
\Cref{tab:ablation_n}~presents the results of varying the number of online rollouts $N$ performed during training. Specifically, we report results for $N=5,~10, ~15$, where $N=10$ is the default setting in our experiments. Interestingly, increasing the rollouts to $N=15$ leads to lower specificity and correctness. 
Similar behavior where smaller group sizes outperform larger ones was reported in a recent study on GRPO~\cite{de2025learning}, which might be due to limitations in batch-based grouping strategies that mix unrelated episodes. With rollouts $N=5$, the model behaves similarly to $N=10$, with minor gain in specificity yet minor drop in correctness, resulting in equal HM values.
\begin{table}[t!]
\caption{\ourmethod with different rollouts size $N$. Best in \textbf{bold}.} 
\vspace{-6pt}
\centering
\small
\setlength{\tabcolsep}{3pt}
\resizebox{\columnwidth}{!}{
\begin{tabular}{lcccccc|ccc}
\toprule
& \multicolumn{6}{c|}{Prediction categorization} & \multicolumn{3}{c}{Metrics}\\
\cmidrule(lr){2-7} \cmidrule(lr){8-10}
N rollouts & $\morespecific$ & $\specific$ & $\lessspecific$ & $\generic$ & $\abstain$ & $\wrong$ & \spe~$\uparrow$ & \cor~$\uparrow$ & HM~$\uparrow$  \\
\midrule
5 & 5.5\% & 63.6\% & 5.8\% & 9.5\% & 0.0\% & 15.6\% & \textbf{0.925} & 0.844 & \textbf{0.883} \\
10 & 5.6\% & 63.4\% & 5.1\% & 10.7\% & 0.0\% & 15.2\% & 0.920 & \textbf{0.848}& \textbf{0.883} \\
15 & 4.6\% & 50.4\% & 3.7\% & 22.2\% & 0.0\% & 19.0\% & 0.848 &  0.810 & 0.824 \\
\bottomrule
\end{tabular}
}
\vspace{-8pt}
\label{tab:ablation_n}
\end{table}



\section{Conclusion}\label{sec:conclusion}
We addressed open-world fine-grained classification with reasoning LMMs, aiming to generate more specific predictions without sacrificing correctness.
Reasoning LMMs are overly generic in recognizing fine-grained visual concepts. Yet, our analysis showed that this is not because they lack domain knowledge, but because they fail to reliably express the most specific prediction they can produce.
We introduced \ourmethod, a specificity-aware reinforcement learning framework that uses a dynamic, sample-wise reward based on the best predictions found in online rollouts. \ourmethod leverages a LLM verifier to provide graded feedback enabling specificity-aware dynamic reward within a GRPO-like policy optimization framework. This design promotes specificity within the model’s inherent capability, preventing the correctness degradation observed in existing approaches.
Out-of-domain comparisons across fine-grained and very fine-grained sets show that \ourmethod consistently achieves the best trade-off between specificity and correctness.

\paragraph{Acknowledgements.}
This work was supported by the Ministero delle Imprese e del Made in Italy (IPCEI Cloud DM 27 giugno 2022 – IPCEI-CL-0000007) and European Union (Next Generation EU).
Additional support was provided by the EU Horizon projects SWARMCHESTRATE (No. 101135012) and ELLIOT (No. 101214398).
The authors acknowledge the CINECA award under the ISCRA initiative for the availability of high-performance computing resources and support.

{
    \small
    \bibliographystyle{ieeenat_fullname}
    \bibliography{main}
}

\clearpage
\appendix
\maketitlesupplementary
In this supplementary material, we present additional details and analyses that complement the content of the main document.
First, in \cref{sec:implementation details}, we provide further implementation details, including the prompts we used for the LMM and the LLM verifier, along with the optimization strategies adopted to improve training efficiency.
Next, in \cref{sec:full_results}, we report the complete out-of-domain evaluation results for each individual dataset in both fine-grained and very fine-grained sets, along with further prompting baselines and additional qualitative examples.
Finally, in \cref{sec:more_ablation}, we extend the ablation studies on the impact of training sets from different domains and the training-set size on the performance of \ourmethod.

\section{Additional implementation details}
\label{sec:implementation details}

\subsection{Prompts}
Here, we report all the prompts used in our experiments. These include the classification prompt $\prompt_c$ provided to the reasoning LMM $\netlmm$, the verification prompt $\prompt_j$ used by the LLM-as-a-judge $\netllm$, and the prompt used to generate the reasoning traces for the supervised fine-tuning (\textit{sft}) baseline.

\subsubsection{LMM prompts}
In our experiments, we consider a total of three different prompts when querying a LMM to classify an image.

\noindent\textbf{Default.} Our default prompt is shown in \cref{fig:vlm}. Since our work focuses on reasoning models, we not only request a classification of the input image, but we also explicitly instruct the model to first perform reasoning and then provide a single label. Specifically, we follow the standard \texttt{<think>}/\texttt{<answer>} tags format.
This structured output simplifies the extraction of the final prediction and its subsequent verification by the LLM-as-a-judge.

\noindent\textbf{``Be specific''.} In the ``Be specific'' baseline, we explicitly encourage the model to be specific in its prediction. To this end, we modify the default prompt by adding the requirement to be specific. The complete text query is reported in \cref{prompt:vlm_specific}.

\noindent\textbf{Format free.} When considering the evaluation protocol in \cite{conti2025large}, for consistency and fair comparison, we adopt the same prompting strategy reported in the original paper~\cite{conti2025large}, as shown in \cref{prompt:ale_eval}. 
Since this previous work does not have a focus on reasoning models, it adopts a more general-purpose prompt without formatting requirements.

\begin{figure}[t]
\begin{tcolorbox}[title=Default LMM prompt ($\prompt_c$), fonttitle=\bfseries,
left=5pt,
right=5pt,
top=3pt,
bottom=3pt,
boxrule=0.5pt, coltitle=black, colbacktitle=gray!20]
Classify the image.

Output the thinking process in \texttt{<think>} \texttt{</think>} and the final answer in \texttt{<answer>} \texttt{</answer>} tags.

The output answer format should be as follows:
\texttt{<think>} ... \texttt{</think>} \texttt{<answer>}a single label or the word `None' to abstain.\texttt{</answer>}.

Please strictly follow the format.
\end{tcolorbox}
\vspace{-3mm}
\caption{LMM default prompt for prediction.}
\label{fig:vlm}
\end{figure}

\begin{figure}[t]
\begin{tcolorbox}[title=``Be specific'' LMM prompt ($\prompt_c$),
left=5pt,
right=5pt,
top=3pt,
bottom=3pt,
fonttitle=\bfseries, boxrule=0.5pt, coltitle=black, colbacktitle=gray!20]
Classify the image, {\color{blue}{be specific}}.

Output the thinking process in \texttt{<think>} \texttt{</think>} and the final answer in \texttt{<answer>} \texttt{</answer>} tags.

The output answer format should be as follows:
\texttt{<think>} ... \texttt{</think>} \texttt{<answer>}a single label or the word `None' to abstain.\texttt{</answer>}.

Please strictly follow the format.
\end{tcolorbox}
\vspace{-3mm}
\caption{LMM prompt for prediction for the ``Be specific'' baseline.}
\label{prompt:vlm_specific}
\end{figure}
\begin{figure}[t]
\begin{tcolorbox}[title=Format free LMM prompt ($\prompt_c$) \cite{conti2025large}, fonttitle=\bfseries,
boxrule=0.5pt,
left=5pt,
right=5pt,
top=3pt,
bottom=3pt,
coltitle=black, colbacktitle=gray!20]
What type of object is in this photo?
\end{tcolorbox}
\vspace{-3mm}
\caption{LMM prompt used in the evaluation protocol of~\cite{conti2025large}.}
\label{prompt:ale_eval}
\end{figure}

\subsubsection{LLM-as-a-judge prompt}
\Cref{prompt:llm_verifier} shows the prompt used when querying the LLM verifier to categorize a prediction into the categories defined in the main paper. 
This prompt provides a precise definition with in-context examples for each category. The placeholder \texttt{\%s} is replaced with the actual \texttt{ground\_truth} and \texttt{prediction} formatted in the specified JSON format.
To eliminate the possibility of invalid responses from the LLM verifier, we utilize the vLLM~\cite{kwon2023efficient} guided decoding strategy to constrain the model in generating only one of the predefined categories as the response.

\begin{figure*}[t]
\centering
\vspace{1cm}
\begin{tcolorbox}[
    title=LLM-as-a-judge prompt ($\prompt_j$),
    fonttitle=\bfseries,
    boxrule=0.5pt,
    coltitle=black,
    colbacktitle=gray!20,
    colback={yellow!20!orange!15}
]

\textbf{Role}: You are an expert AI classifier. Your goal is to classify a model's \texttt{prediction} against a \texttt{ground\_truth} label.

\textbf{Task}: You will receive a single JSON object. Your output must be \textbf{only the classification category} and nothing else.

\medskip
\hrule
\medskip

\textbf{Classification Categories}

\begin{itemize}

    \item \textbf{Specific}: The prediction is an exact match or a direct synonym for the ground truth. This includes common name/scientific name equivalence.\\[2pt]
    \begin{tabular}{@{}l l@{}}
    \texttt{prediction: "Panthera leo"} & \texttt{ground\_truth: "lion"} \\
    \texttt{prediction: "passiflora"}   & \texttt{ground\_truth: "passion flower"} \\
    \end{tabular}

    \item \textbf{Less Specific}: The prediction is a correct, but \textbf{closely related parent category} (e.g., family, genus, product line) of the ground truth.\\[2pt]
    \begin{tabular}{@{}l l@{}}
    \texttt{prediction: "Warbler"}     & \texttt{ground\_truth: "Golden-winged Warbler"} \\
    \texttt{prediction: "Boeing 707"}  & \texttt{ground\_truth: "707-320"} \\
    \end{tabular}

    \item \textbf{Generic}: The prediction is correct, but a \textbf{significantly broader category} than the ground truth.\\[2pt]
    \begin{tabular}{@{}l l@{}}
    \texttt{prediction: "dog"}               & \texttt{ground\_truth: "samoyed"} \\
    \texttt{prediction: "Commercial Airline"} & \texttt{ground\_truth: "757-200"} \\
    \end{tabular}

    \item \textbf{More Specific}: The prediction is a correct, but \textbf{more specific subtype or instance} of the ground truth.\\[2pt]
    \begin{tabular}{@{}l l@{}}
    \texttt{prediction: "samoyed"}  & \texttt{ground\_truth: "dog"} \\
    \texttt{prediction: "757-200"}  & \texttt{ground\_truth: "Commercial Airline"} \\
    \end{tabular}

    \item \textbf{Wrong}: The prediction is factually incorrect, contradictory, malformed, completely unrelated to the ground truth, or contains multiple options.\\[2pt]
    \begin{tabular}{@{}l l@{}}
    \texttt{prediction: "cat"}                   & \texttt{ground\_truth: "dog"} \\
    \texttt{prediction: "Blue-winged Warbler"}   & \texttt{ground\_truth: "Golden-winged Warbler"} \\
    \texttt{prediction: "b1rd"}                  & \texttt{ground\_truth: "bird"} \\
    \texttt{prediction: "robin or cardinal"}     & \texttt{ground\_truth: "bird"} \\
    \texttt{prediction: "\_prototype"}           & \texttt{ground\_truth: "Boeing 717"} \\
    \end{tabular}

    \item \textbf{Abstain}: The prediction is a refusal to answer.\\[2pt]
    \begin{tabular}{@{}l l@{}}
    \texttt{prediction: "none"}        & \\
    \texttt{prediction: "I don't know"} & \\
    \texttt{prediction: "Cannot tell"}  & \\
    \end{tabular}

\end{itemize}

\medskip
\textbf{Input Format}:  
You will receive a single JSON object with the following structure:
\begin{verbatim}
{"ground_truth": "<the_ground_truth_label>", 
 "prediction": "<the_vlm_prediction>"}
\end{verbatim}

\textbf{Output Format}:  
Your response must be a \textbf{single word} representing the classification category.

\medskip
\textbf{Prompt}:
\begin{verbatim}
Classify the prediction in the following JSON object based on 
the rules provided. Your output must be a single word.

INPUT:
%s
\end{verbatim}

\end{tcolorbox}
\caption{Prompt for the LLM-as-a-judge verifier categorizing a prediction given the target ground-truth.}
\label{prompt:llm_verifier}
\end{figure*}

\subsubsection{CoT generation prompt}
\Cref{prompt:cot_generation} reports the prompt used to generate a chain-of-thought reasoning trace for each sample in the training set, which are then used to construct the custom dataset for supervised fine-tuning.
This prompt provides the LMM with the ground-truth label associated to the image, and requests a thinking trace leading to the correct prediction. 
\begin{figure*}[t]
\begin{tcolorbox}[
    title=CoT generation prompt,
    fonttitle=\bfseries,
    boxrule=0.5pt,
    coltitle=black,
    colbacktitle=gray!20,
    colback={blue!10}
]

Given the image and the correct classification label: \{ground\_truth\}.

Generate a correct well-reasoned response that will answer the following question:

\begin{quote}
Classify the image.\\
Output the thinking process in \texttt{<think>} \texttt{</think>} and the final answer in \texttt{<answer>} \texttt{</answer>} tags.\\
The output answer format should be as follows: 
\texttt{<think>} \ldots \texttt{</think>} \texttt{<answer>}a single label or the word 'None' to abstain.\texttt{</answer>}.\\
Please strictly follow the format.
\end{quote}

Describe the content of the image, then infer the correct classification label.  
The thinking process must proceed without assuming or referencing the true label in advance.  
Use the correct classification label in the final answer and strictly follow the format.

\end{tcolorbox}
\vspace{-3mm}
\caption{Prompt for generating the reasoning traces used to train the supervised fine-tuning baseline model.}
\label{prompt:cot_generation}
\end{figure*}

\subsection{Optimizations}
Our study can be computationally demanding at training and evaluation due to the LMM inference and LLM-as-a-judge evaluation. 
We therefore adopt several optimizations strategies to reduce computational costs.

\noindent \textbf{Inference Engine.}
In our experiments, we used the vLLM~\cite{kwon2023efficient} inference engine both to generate the LMM predictions and to compute the LLM-as-a-judge categorization. This engine is highly optimized and enabled a significant speed-up of the evaluation process. 
Among its key features, it includes PagedAttention~\cite{kwon2023efficient} for efficient memory management, continuous batching, which is crucial in our setting where variable-size image inputs make static batch selection difficult, and prefix caching, which is beneficial since our textual prompt is mostly fixed.
For instance, generating 1000 predictions for Flowers102 with Qwen2.5-VL-7B on a A100 64 GB GPU takes 2.27 minutes with vLLM. In comparison, a naive PyTorch implementation requires 25.11 minutes, using a batch size of 32, which is the largest batch size avoiding out-of-memory errors across all our evaluation datasets. The PyTorch implementation incurs computation time that is a magnitude higher than using vLLM.
Only when following the evaluation protocol in \cite{conti2025large}, we used the same testing code provided by the authors, which is built on PyTorch.

\begin{figure}
    \centering
    \includegraphics[width=\columnwidth]{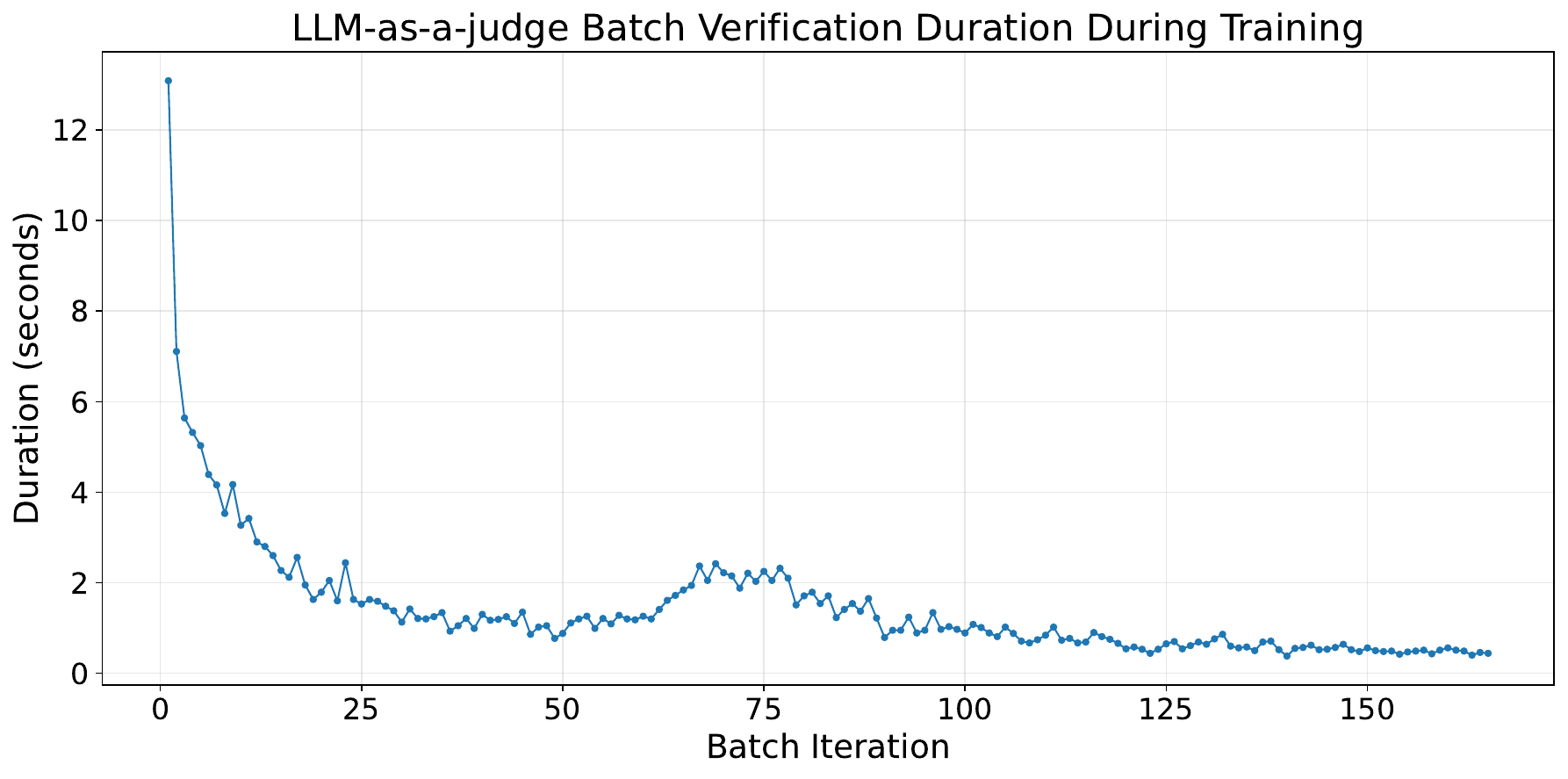}
    \caption{LLM-as-a-judge per-batch verification times during reinforcement learning training, showing the speedup obtained as cache hit rates increase when starting from an empty cache.}
    \label{fig:batch_duration}
\end{figure}

\noindent \textbf{LLM-as-a-judge optimization via caching.}
We implemented a caching mechanism to reduce the verification time of the LLM-as-a-judge categorization procedure. 
This system stores a dictionary where (prediction, ground\_truth) pairs are associated to the corresponding verification\_category. This avoids repeating the LLM verification of a pair that has already been categorized in a previous computation. The cached data is persistent, allowing results to be reused across different runs.
We used this cache-based solution to speed up the categorization process both during evaluation and during the reward computation in RL training. 
During evaluation, we run Llama-3-72B~\cite{dubey2024llama3} using vLLM with tensor parallelism set to 4, distributing the model across four A100 GPUs. 
For a test subsample of 1000 predictions from Flowers102, our optimized implementation, with an initially empty cache, completes verification in 6.77 seconds, with only 301 actual LLM calls and a 70\% cache hit rate. 
During reinforcement learning, we use a total of six A100 GPUs: one four-GPU node running the training loop with verl (an open source implementation of ~\cite{sheng2024hybridflow}) and two additional GPUs on a separate node performing batched LLM-as-a-judge inference using Qwen3-30B-A3B-Instruct-2507-FP8~\cite{qwen3technicalreport} with tensor parallelism set to 2. 
With a batch size of 256 and 10 rollouts, each verification batch contains 2560 predictions. Analyzing the reward calculation durations shown in  \cref{fig:batch_duration}, we see an initial warm-up phase in which early batches require 2-14 seconds while the cache is being populated. Afterwards, the processing time quickly drops and stabilizes at approximately 0.5 to 1 seconds per batch, except for a mid-training bump that may be caused by cache misses caused by the model exploration. Overall, reinforcement learning training takes approximately 12 hours using our optimized implementation.

\begin{figure*}
    \centering
    \includegraphics[width=\textwidth]{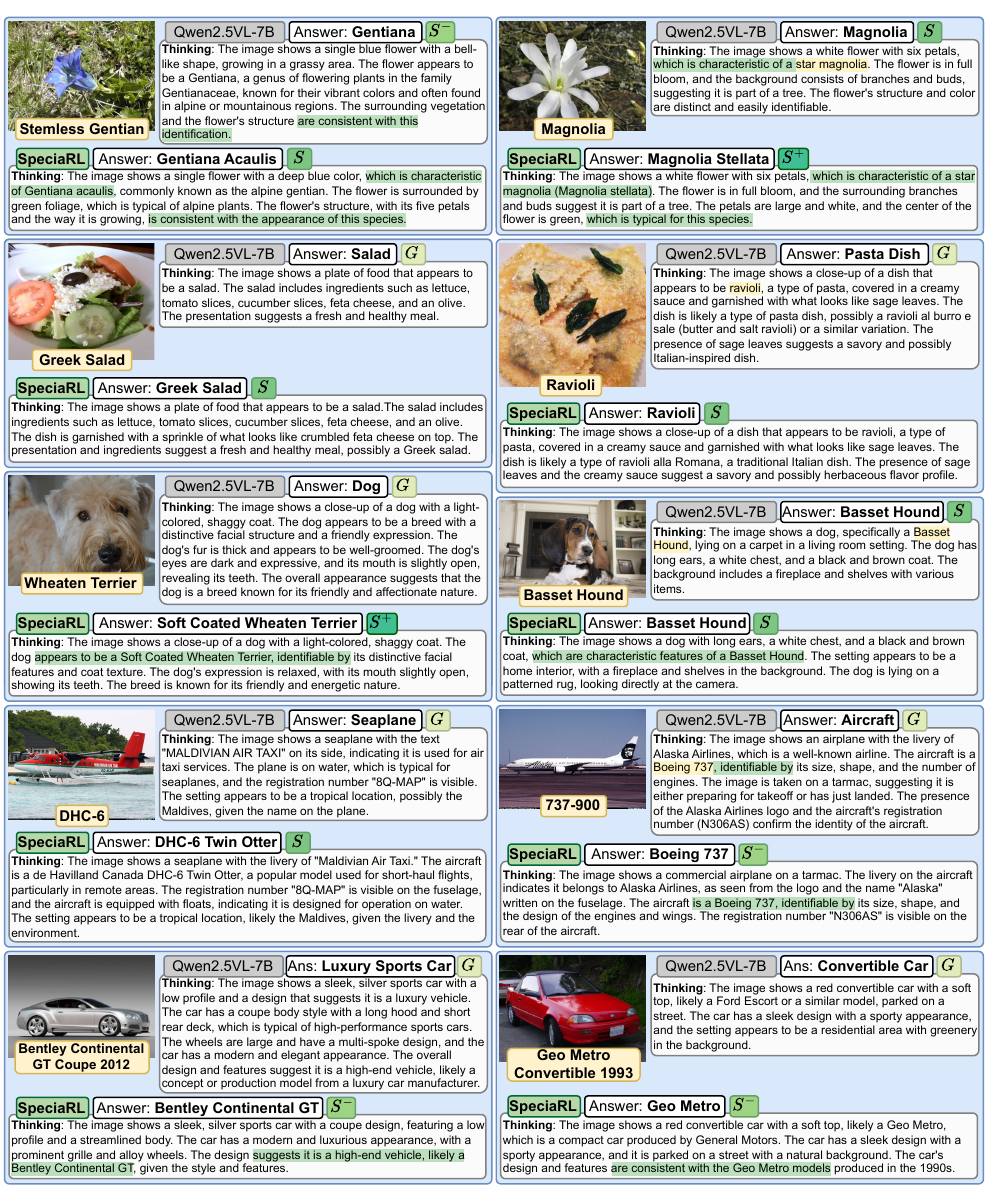}
    \caption{Additional qualitative examples of the think-answer output of the base model Qwen2.5VL-7B and \ourmethod.}
    \label{fig:qualitatives supmat}
\end{figure*}

\begin{figure}
    \centering
    \includegraphics[width=\columnwidth]{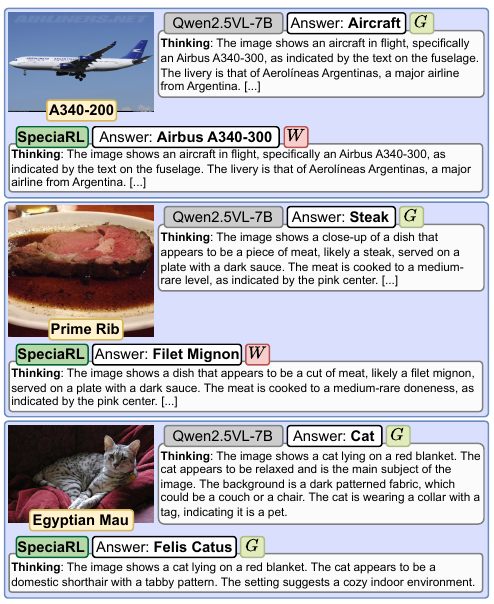}
    \caption{\textbf{Failure cases.} Qualitative examples of \ourmethod providing a Wrong prediction (Top \& Center) and of \ourmethod unnecessarily using a scientific name for a generic concept (Bottom).}
    \label{fig:failures}
\end{figure}

\begin{table*}[ht]
\centering
\footnotesize
\setlength{\tabcolsep}{4pt}
\begin{tabular}{l lcccccc|ccc}
\toprule
\multicolumn{11}{c}{\cellcolor{gray!15}\textbf{Fine-grained}} \\
\midrule
 & & \multicolumn{6}{c|}{Prediction categorization} & \multicolumn{3}{c}{Metrics} \\
\cmidrule(lr){3-8} \cmidrule(lr){9-11}
Dataset & Model & $\morespecific$ & $\specific$ & $\lessspecific$ & $\generic$ & $\abstain$ & $\wrong$ & \spe~$\uparrow$ & \cor~$\uparrow$ & HM~$\uparrow$ \\
\midrule
\multirow{11}{*}{Flowers102~\cite{nilsback2008flowers}} & \cellcolor{lightblue}CaSED~\cite{conti2023vocabulary} & 0.0\% & 57.4\% & 8.4\% & 14.7\% & 0.0\% & 19.4\% & 0.883 & 0.806 & 0.842 \\
 & \cellcolor{lightblue}InternVL2.5-4B~\cite{chen2024expanding} & 0.2\% & 15.8\% & 1.8\% & 29.8\% & 20.0\% & 32.4\% & 0.551 & 0.676 & 0.607 \\
 & \cellcolor{lightblue}InternVL2.5-8B~\cite{chen2024expanding} & 0.4\% & 26.4\% & 3.3\% & 15.1\% & 13.2\% & 41.6\% & 0.688 & 0.584 & 0.632 \\
 & \cellcolor{lightblue}Qwen2.5VL-3B~\cite{bai2025qwen25} & 0.1\% & 26.6\% & 2.7\% & 49.6\% & 1.9\% & 19.2\% & 0.668 & 0.808 & 0.731 \\
 & \cellcolor{lightblue}Qwen2.5VL-7B~\cite{bai2025qwen25} & 0.1\% & 47.2\% & 4.1\% & 34.8\% & 1.2\% & 12.7\% & 0.779 & 0.873 & 0.823 \\
 & \cellcolor{lightblue}Qwen2.5VL-7B (``Be specific'') & 0.2\% & 63.5\% & 5.8\% & 12.7\% & 3.0\% & 14.7\% & 0.882 & 0.853 & 0.867 \\
\cmidrule(lr){2-11}
 & \cellcolor{lightgreen}Qwen2.5VL-7B (\textit{sft}) & 1.3\% & 69.6\% & 8.5\% & 3.0\% & 0.0\% & 17.5\% & 0.956 & 0.825 & 0.885 \\
 & \cellcolor{lightgreen}Qwen2.5VL-7B (\textit{rft}) & 10.4\% & 70.3\% & 5.4\% & 1.5\% & 0.0\% & 12.4\% & \underline{0.976} & \underline{0.876} & \underline{0.923} \\
\cmidrule(lr){2-11}
 & \cellcolor{lightgreen}\textbf{\ourmethod-7B} & 13.6\% & 69.2\% & 5.0\% & 1.7\% & 0.0\% & 10.5\% & \textbf{0.976} & \textbf{0.895} & \textbf{0.934} \\
\cmidrule(lr){2-11}
 & \cellcolor{gray!10}\textit{Qwen2.5VL-7B (BoN-64)} & 4.4\% & 78.3\% & 3.7\% & 9.9\% & 0.6\% & 3.1\% & 0.935 & 0.969 & 0.952 \\
\midrule
\multirow{11}{*}{Food101~\cite{bossard2014food}} & \cellcolor{lightblue}CaSED~\cite{conti2023vocabulary} & 0.0\% & 33.0\% & 13.2\% & 35.3\% & 0.0\% & 18.5\% & 0.743 & 0.815 & 0.777 \\
 & \cellcolor{lightblue}InternVL2.5-4B~\cite{chen2024expanding} & 0.5\% & 10.5\% & 1.4\% & 71.4\% & 2.6\% & 13.7\% & 0.560 & 0.863 & 0.680 \\
 & \cellcolor{lightblue}InternVL2.5-8B~\cite{chen2024expanding} & 0.8\% & 10.6\% & 1.5\% & 46.3\% & 30.2\% & 10.7\% & 0.483 & \textbf{0.893} & 0.627 \\
 & \cellcolor{lightblue}Qwen2.5VL-3B~\cite{bai2025qwen25} & 1.5\% & 17.9\% & 2.5\% & 53.4\% & 7.8\% & 16.9\% & 0.601 & 0.831 & 0.697 \\
 & \cellcolor{lightblue}Qwen2.5VL-7B~\cite{bai2025qwen25} & 1.3\% & 32.0\% & 3.8\% & 47.8\% & 2.0\% & 13.2\% & 0.697 & \underline{0.868} & 0.773 \\
 & \cellcolor{lightblue}Qwen2.5VL-7B (``Be specific'') & 1.8\% & 38.0\% & 4.6\% & 34.7\% & 5.6\% & 15.3\% & 0.732 & 0.847 & 0.785 \\
\cmidrule(lr){2-11}
 & \cellcolor{lightgreen}Qwen2.5VL-7B (\textit{sft}) & 3.5\% & 51.4\% & 9.1\% & 11.6\% & 0.5\% & 24.0\% & \underline{0.889} & 0.760 & \underline{0.820} \\
 & \cellcolor{lightgreen}Qwen2.5VL-7B (\textit{rft}) & 3.2\% & 52.0\% & 7.4\% & 8.7\% & 0.1\% & 28.6\% & \textbf{0.912} & 0.714 & 0.801 \\
\cmidrule(lr){2-11}
 & \cellcolor{lightgreen}\textbf{\ourmethod-7B} & 1.2\% & 54.3\% & 5.8\% & 19.7\% & 0.0\% & 18.9\% & 0.860 & 0.811 & \textbf{0.835} \\
\cmidrule(lr){2-11}
 & \cellcolor{gray!10}\textit{Qwen2.5VL-7B (BoN-64)} & 15.1\% & 52.1\% & 5.5\% & 26.5\% & 0.5\% & 0.2\% & 0.849 & 0.998 & 0.917 \\
\midrule
\multirow{11}{*}{OxfordPets~\cite{parkhi2012pets}} & \cellcolor{lightblue}CaSED~\cite{conti2023vocabulary} & 0.0\% & 40.7\% & 10.2\% & 22.5\% & 0.0\% & 26.5\% & 0.812 & 0.735 & 0.772 \\
 & \cellcolor{lightblue}InternVL2.5-4B~\cite{chen2024expanding} & 0.1\% & 7.9\% & 1.3\% & 61.8\% & 2.6\% & 26.2\% & 0.550 & 0.738 & 0.630 \\
 & \cellcolor{lightblue}InternVL2.5-8B~\cite{chen2024expanding} & 0.9\% & 13.2\% & 5.2\% & 30.6\% & 18.7\% & 31.5\% & 0.554 & 0.685 & 0.613 \\
 & \cellcolor{lightblue}Qwen2.5VL-3B~\cite{bai2025qwen25} & 0.8\% & 7.4\% & 2.9\% & 57.3\% & 3.1\% & 28.6\% & 0.557 & 0.714 & 0.626 \\
 & \cellcolor{lightblue}Qwen2.5VL-7B~\cite{bai2025qwen25} & 2.7\% & 35.1\% & 5.2\% & 35.6\% & 1.0\% & 20.4\% & 0.751 & 0.796 & 0.773 \\
 & \cellcolor{lightblue}Qwen2.5VL-7B (``Be specific'') & 4.3\% & 45.8\% & 8.2\% & 19.7\% & 1.6\% & 20.4\% & 0.835 & 0.796 & 0.815 \\
\cmidrule(lr){2-11}
 & \cellcolor{lightgreen}Qwen2.5VL-7B (\textit{sft}) & 2.4\% & 72.1\% & 5.3\% & 3.2\% & 0.4\% & 16.5\% & \textbf{0.961} & \underline{0.835} & \textbf{0.894} \\
 & \cellcolor{lightgreen}Qwen2.5VL-7B (\textit{rft}) & 0.3\% & 34.7\% & 2.1\% & 39.0\% & 0.0\% & 23.8\% & 0.737 & 0.762 & 0.749 \\
\cmidrule(lr){2-11}
 & \cellcolor{lightgreen}\textbf{\ourmethod-7B} & 2.1\% & 66.6\% & 4.5\% & 10.7\% & 0.0\% & 16.1\% & \underline{0.923} & \textbf{0.839} & \underline{0.879} \\
\cmidrule(lr){2-11}
 & \cellcolor{gray!10}\textit{Qwen2.5VL-7B (BoN-64)} & 12.9\% & 59.8\% & 5.8\% & 19.5\% & 0.5\% & 1.4\% & 0.882 & 0.986 & 0.931 \\
\bottomrule
\end{tabular}
\caption{Results on the individual datasets composing the fine-grained set.}
\label{tab:complete_fine-grained}
\end{table*}
\begin{table*}[ht]
\centering
\footnotesize
\setlength{\tabcolsep}{4pt}
\begin{tabular}{l lcccccc|ccc}
\toprule
\multicolumn{11}{c}{\cellcolor{gray!15}\textbf{Very fine-grained}} \\
\midrule
 & & \multicolumn{6}{c|}{Prediction categorization} & \multicolumn{3}{c}{Metrics} \\
\cmidrule(lr){3-8} \cmidrule(lr){9-11}
Dataset & Model & $\morespecific$ & $\specific$ & $\lessspecific$ & $\generic$ & $\abstain$ & $\wrong$ & \spe~$\uparrow$ & \cor~$\uparrow$ & HM~$\uparrow$ \\
\midrule
\multirow{11}{*}{FGVCAircraft~\cite{maji2013aircraft}} & \cellcolor{lightblue}CaSED~\cite{conti2023vocabulary} & 0.0\% & 1.6\% & 13.9\% & 37.7\% & 0.0\% & 46.8\% & 0.580 & 0.532 & 0.555 \\
 & \cellcolor{lightblue}InternVL2.5-4B~\cite{chen2024expanding} & 0.0\% & 0.0\% & 0.2\% & 66.0\% & 8.5\% & 25.3\% & 0.472 & 0.747 & 0.579 \\
 & \cellcolor{lightblue}InternVL2.5-8B~\cite{chen2024expanding} & 0.1\% & 2.2\% & 1.3\% & 59.1\% & 13.0\% & 24.4\% & 0.476 & 0.756 & 0.584 \\
 & \cellcolor{lightblue}Qwen2.5VL-3B~\cite{bai2025qwen25} & 0.2\% & 1.6\% & 1.4\% & 82.4\% & 0.3\% & 14.1\% & 0.514 & 0.859 & 0.643 \\
 & \cellcolor{lightblue}Qwen2.5VL-7B~\cite{bai2025qwen25} & 0.1\% & 6.6\% & 5.4\% & 80.7\% & 0.5\% & 6.7\% & 0.549 & \textbf{0.933} & 0.691 \\
 & \cellcolor{lightblue}Qwen2.5VL-7B (``Be specific'') & 0.5\% & 23.0\% & 20.8\% & 40.4\% & 1.2\% & 14.0\% & 0.693 & \underline{0.860} & 0.768 \\
\cmidrule(lr){2-11}
 & \cellcolor{lightgreen}Qwen2.5VL-7B (\textit{sft}) & 1.0\% & 42.9\% & 33.4\% & 2.3\% & 0.1\% & 20.2\% & 0.879 & 0.798 & \underline{0.837} \\
 & \cellcolor{lightgreen}Qwen2.5VL-7B (\textit{rft}) & 2.2\% & 45.9\% & 25.0\% & 2.0\% & 0.0\% & 25.0\% & \textbf{0.904} & 0.750 & 0.820 \\
\cmidrule(lr){2-11}
 & \cellcolor{lightgreen}\textbf{\ourmethod-7B} & 1.9\% & 46.5\% & 29.0\% & 1.7\% & 0.0\% & 20.9\% & \underline{0.897} & 0.791 & \textbf{0.841} \\
\cmidrule(lr){2-11}
 & \cellcolor{gray!10}\textit{Qwen2.5VL-7B (BoN-64)} & 3.4\% & 48.9\% & 24.6\% & 22.9\% & 0.1\% & 0.1\% & 0.823 & 0.999 & 0.903 \\
\midrule
\multirow{11}{*}{StanfordCars~\cite{krause20133cars}} & \cellcolor{lightblue}CaSED~\cite{conti2023vocabulary} & 0.0\% & 0.2\% & 13.7\% & 74.3\% & 0.0\% & 11.8\% & 0.540 & 0.882 & 0.669 \\
 & \cellcolor{lightblue}InternVL2.5-4B~\cite{chen2024expanding} & 0.0\% & 0.1\% & 2.3\% & 59.4\% & 2.6\% & 35.6\% & 0.499 & 0.644 & 0.563 \\
 & \cellcolor{lightblue}InternVL2.5-8B~\cite{chen2024expanding} & 0.0\% & 0.2\% & 10.2\% & 50.0\% & 18.2\% & 21.4\% & 0.476 & 0.786 & 0.593 \\
 & \cellcolor{lightblue}Qwen2.5VL-3B~\cite{bai2025qwen25} & 0.0\% & 0.5\% & 6.3\% & 67.8\% & 4.6\% & 20.8\% & 0.509 & 0.792 & 0.619 \\
 & \cellcolor{lightblue}Qwen2.5VL-7B~\cite{bai2025qwen25} & 0.0\% & 1.3\% & 20.1\% & 68.4\% & 0.8\% & 9.4\% & 0.561 & 0.906 & 0.693 \\
 & \cellcolor{lightblue}Qwen2.5VL-7B (``Be specific'') & 0.1\% & 2.1\% & 37.8\% & 50.8\% & 1.3\% & 8.0\% & 0.611 & \textbf{0.920} & 0.734 \\
\cmidrule(lr){2-11}
 & \cellcolor{lightgreen}Qwen2.5VL-7B (\textit{sft}) & 0.0\% & 2.1\% & 68.1\% & 21.2\% & 0.1\% & 8.4\% & 0.698 & 0.916 & 0.792 \\
 & \cellcolor{lightgreen}Qwen2.5VL-7B (\textit{rft}) & 0.2\% & 3.5\% & 82.8\% & 5.0\% & 0.0\% & 8.5\% & \textbf{0.746} & 0.915 & \textbf{0.822} \\
\cmidrule(lr){2-11}
 & \cellcolor{lightgreen}\textbf{\ourmethod-7B} & 0.2\% & 3.8\% & 79.4\% & 8.4\% & 0.0\% & 8.2\% & \underline{0.738} & \underline{0.918} & \underline{0.818} \\
\cmidrule(lr){2-11}
 & \cellcolor{gray!10}\textit{Qwen2.5VL-7B (BoN-64)} & 0.5\% & 12.4\% & 60.6\% & 26.3\% & 0.0\% & 0.3\% & 0.716 & 0.997 & 0.834 \\
\bottomrule
\end{tabular}
\caption{Individual dataset results on the very fine-grained set.}
\label{tab:complete_veryfine-grained}
\end{table*}

\section{Additional experimental analysis}
\label{sec:full_results}

We provide per-dataset evaluations of our method, in-domain evaluation on the CUB~\cite{wah2011caltech} test set, additional qualitative examples, additional results for prompting-based baselines, and extended ablation studies.

\subsection{Per-dataset evaluation}
In the main paper, we reported results averaged over the \textit{\textbf{fine-grained}} and the \textbf{\textit{very fine-grained}} test sets. Here, we present the results for each individual dataset, with \cref{tab:complete_fine-grained} corresponding to the fine-grained ones and \cref{tab:complete_veryfine-grained} to the very fine-grained ones.
Considering overall performance, measured by the harmonic mean (HM), our \ourmethod achieves the best performance on three out of five benchmarks (Flowers102, Food101, FGVAircraft) and the second best on the remaining two (OxfordPets, StanfordCars).
Notably, on three datasets (Flowers102, OxfordPets, StanfordCars), our method not only improves specificity relatively to the base model, but also correctness.
Overall, \ourmethod performs strongly on all evaluation benchmarks, even though these datasets span domains significantly different from CUB~\cite{wah2011caltech}, which is used for training. These results support the effectiveness of our method in eliciting a general classification behavior oriented towards both specificity and correctness.

\subsection{In-domain evaluation}

\begin{table}[t!]
\caption{In-domain evaluation of the training strategies.} 
\vspace{-6pt}
\centering
\small
\setlength{\tabcolsep}{3pt}
\resizebox{\columnwidth}{!}{
\begin{tabular}{l lcccccc|ccc}
\toprule
\multicolumn{11}{c}{\cellcolor{gray!15}\textbf{In-domain}} \\
\midrule
 & & \multicolumn{6}{c|}{Prediction categorization} & \multicolumn{3}{c}{Metrics} \\
\cmidrule(lr){3-8} \cmidrule(lr){9-11}
Dataset & Model & $\morespecific$ & $\specific$ & $\lessspecific$ & $\generic$ & $\abstain$ & $\wrong$ & \spe~$\uparrow$ & \cor~$\uparrow$ & HM~$\uparrow$ \\
\midrule
\multirow{6}{*}{CUB~\cite{wah2011caltech}} & \cellcolor{lightblue}Qwen2.5VL-7B~\cite{bai2025qwen25} & 0.2\% & 23.0\% & 15.9\% & 48.1\% & 2.0\% & 11.0\% & 0.669 & 0.890 & 0.764 \\
& \cellcolor{lightblue}Qwen2.5VL-7B (``Be specific'') & 0.2\% & 32.2\% & 13.7\% & 35.3\% & 2.6\% & 16.1\% & 0.726 & 0.839 & 0.779 \\

\cmidrule(lr){2-11}
 & \cellcolor{lightgreen}Qwen2.5VL-7B (\textit{sft}) & 0.1\% & 80.4\% & 0.7\% & 0.3\% & 0.0\% & 18.5\% & 0.996 & 0.815 & 0.896 \\
 & \cellcolor{lightgreen}Qwen2.5VL-7B (\textit{rft}) & 1.0\% & 92.7\% & 0.0\% & 0.0\% & 0.0\% & 6.3\% & \textbf{1.000} & \textbf{0.937} & \textbf{0.968} \\
\cmidrule(lr){2-11}
 & \cellcolor{lightgreen}\textbf{\ourmethod-7B} & 0.6\% & 92.7\% & 0.0\% & 0.0\% & 0.0\% & 6.7\% & \textbf{1.000} & \underline{0.933} & \underline{0.965} \\
\cmidrule(lr){2-11}
 & \cellcolor{gray!10}\textit{Qwen2.5VL-7B (BoN-64)} & 1.1\% & 58.0\% & 14.1\% & 26.4\% & 0.1\% & 0.3\% & 0.831 & 0.997 & 0.907 \\
\bottomrule
\end{tabular}
}
\vspace{-8pt}
\label{tab:half_indomain_result}
\end{table}

The fine-tuned models in the main results are trained on the same subset of CUB~\cite{wah2011caltech}, implying that evaluations on the fine-grained and very fine-grained sets are out-of-domain. \cref{tab:half_indomain_result} reports the in-domain performance on the CUB test-split.
In this setting, all training-based variants achieve very high specificity, exceeding BoN-64. In terms of correctness, only the RL-based methods improve over the base model, although they remain below BoN-64. Overall, the best harmonic mean is obtained by the two RL-based approaches, surpassing BoN-64. These results suggests that the models not only learn to adjust their classification behavior to be
more specific and correct, but also acquire domain-specific
information. Importantly, this does hinder generalization, as
demonstrated by the strong out-of-domain performance in
our extensive evaluations.

\subsection{Additional qualitative results}
We showcase additional qualitative classification outputs, two per test dataset, in \cref{fig:qualitatives supmat}. Examples in the same row are sampled from the same dataset, ordered from top to bottom as follows: Flowers102~\cite{nilsback2008flowers}, Food101~\cite{bossard2014food}, OxfordPets~\cite{parkhi2012pets}, FGVCAircraft~\cite{maji2013aircraft}, and StanfordCars~\cite{krause20133cars}.
In line with our quantitative evaluation, our \ourmethod consistently produces more specific classifications than the base model Qwen2.5VL-7B. The reasoning traces of \ourmethod contain frequent reference to fine-grained visual evidences that support the final prediction or the intermediate reasoning process (highlighted in green). The base model (Qwen2.5VL-7B) exhibits such behavior more rarely.
Interestingly, we observe cases where the base model identifies a more specific label during the reasoning process (highlighted in yellow), yet outputs a more generic label as the final prediction.
This observation further supports our hypothesis that the base model does possess the knowledge and reasoning capabilities to be more precise, however it is biased towards more generic predictions.

We also investigate failure cases of \ourmethod and report some qualitative examples in \cref{fig:failures}.
Although our training strategy aims to increase specificity without sacrificing correctness, we find instances where our \ourmethod makes Wrong predictions when attempting to be specific in its classification (see Top \& Center examples in the figure).
Also, we notice that \ourmethod sometimes uses scientific names even when referring to generic concepts. For example, we found it predicts \lab{``Felis Catus''} or \lab{``Canis Lupus Familiaris''} instead of \lab{``Cat''} or \lab{``Dog''} (see Bottom example in the figure). While these predictions are unusual, the LLM verifier correctly categorizes them as Generic.
We hypothesize that this interesting behavior could be inherited from training on the CUB~\cite{wah2011caltech} bird-species dataset, where the model is positively rewarded for specific scientific names.

\begin{figure}[t]
\vspace{6mm}
\begin{tcolorbox}[title=Additional LMM prompt ($\prompt_c$ (v1)), fonttitle=\bfseries,
left=5pt,
right=5pt,
top=3pt,
bottom=3pt,
boxrule=0.5pt, coltitle=black, colbacktitle=gray!20]
Classify the image.

Prioritize correctness first. Be as specific as you can ONLY when you are confident the finer-grained label is correct. If you are not confident about a fine-grained label, output a more general but correct label instead. If you cannot provide a correct label, output 'None'.

Output the thinking process in \texttt{<think>} \texttt{</think>} and the final answer in \texttt{<answer>} \texttt{</answer>} tags.

The output answer format should be as follows:
\texttt{<think>} ... \texttt{</think>} \texttt{<answer>}a single label or the word `None' to abstain.\texttt{</answer>}.

Please strictly follow the format.
\end{tcolorbox}
\vspace{-3mm}
\caption{Generated LMM prompt ($\prompt_c$ (v1)).}
\label{fig:vlmv1}
\end{figure}
\begin{figure}[t]
\begin{tcolorbox}[title=Additional LMM prompt ($\prompt_c$ (v2)), fonttitle=\bfseries,
left=5pt,
right=5pt,
top=3pt,
bottom=3pt,
boxrule=0.5pt, coltitle=black, colbacktitle=gray!20]
Classify the image.

Optimize for high precision: do not guess. If you are unsure, abstain with 'None'. Only output a label when you can justify it from clear visual evidence in the image. When you do output a label, make it the most specific label that the evidence supports.

Output the thinking process in \texttt{<think>} \texttt{</think>} and the final answer in \texttt{<answer>} \texttt{</answer>} tags.

The output answer format should be as follows:
\texttt{<think>} ... \texttt{</think>} \texttt{<answer>}a single label or the word `None' to abstain.\texttt{</answer>}.

Please strictly follow the format.
\end{tcolorbox}
\vspace{-3mm}
\caption{Generated LMM prompt ($\prompt_c$ (v2)).}
\label{fig:vlmv2}
\end{figure}
\begin{figure}[t]
\begin{tcolorbox}[title=Additional LMM prompt ($\prompt_c$ (v3)), fonttitle=\bfseries,
left=5pt,
right=5pt,
top=3pt,
bottom=3pt,
boxrule=0.5pt, coltitle=black, colbacktitle=gray!20]
Classify the image.

Only care about precision/specificity: always output the most fine-grained label you can. Do not abstain. Do not output 'None'. If multiple fine-grained labels are plausible, choose the single most specific label you consider most likely.

Output the thinking process in \texttt{<think>} \texttt{</think>} and the final answer in \texttt{<answer>} \texttt{</answer>} tags.

The output answer format should be as follows:
\texttt{<think>} ... \texttt{</think>} \texttt{<answer>}a single label or the word `None' to abstain.\texttt{</answer>}.

Please strictly follow the format.
\end{tcolorbox}
\caption{Generated LMM prompt ($\prompt_c$ (v3)).}
\label{fig:vlmv3}
\end{figure}
\begin{table}[h!]
\centering
\small
\resizebox{\columnwidth}{!}{%
    \begin{tabular}{lccc|ccc}
    \toprule
    & \multicolumn{3}{c|}{\cellcolor{gray!15}\textbf{Fine-grained}} & \multicolumn{3}{c}{\cellcolor{gray!15}\textbf{Very fine-grained}} \\
    \cmidrule(lr){2-4} \cmidrule(lr){5-7}
     & \spe~$\uparrow$ & \cor~$\uparrow$ & HM~$\uparrow$ & \spe~$\uparrow$ & \cor~$\uparrow$ & HM~$\uparrow$\\
     \midrule
    $\prompt_c$ (``Be specific'')  & 0.816 & 0.832 & 0.822  & 0.652 & 0.89 & 0.751\\
    $\prompt_c$ (v1)  & 0.840 & 0.830 & 0.834 & 0.688 & 0.885 & 0.772 \\
    $\prompt_c$ (v2) & 0.814 & 0.849 & 0.830 & 0.637 & 0.902 & 0.746 \\
    $\prompt_c$ (v3) & 0.884 & 0.764 & 0.819 & 0.777 & 0.832 & 0.799 \\
    \bottomrule
    \end{tabular}
    \label{tab:data_additional_prompt}
}
\caption{Performance comparison of additional prompting baseline.}
\label{tab:prompting_baseline}
\vspace{-4mm}
\end{table}

\subsection{Additional Prompting baselines}
We report the performance of three additional top-performing variants of the $\prompt_c$ prompt. These variants were generated using ChatGPT by requesting three different optimal predictor prompts given the full task context. As shown in ~\cref{tab:prompting_baseline}, while performance varies across prompt designs, the overall impact is less significant compared to the gains achieved by the training-based methods reported in the main paper. The full text for these variants is provided in Prompts~\ref{fig:vlmv1}, \ref{fig:vlmv2}, and \ref{fig:vlmv3}.

\begin{table*}[t]
\centering
\footnotesize
\setlength{\tabcolsep}{4pt}
\begin{tabular}{l lcccccc|ccc}
\toprule
\multicolumn{11}{c}{\cellcolor{gray!15}\textbf{Fine-grained}} \\
\midrule
 & & \multicolumn{6}{c|}{Prediction categorization} & 
    \multicolumn{3}{c}{Metrics} \\
\cmidrule(lr){3-8} \cmidrule(lr){9-11}
Test set & Training set & $\morespecific$ & $\specific$ &
$\lessspecific$ & $\generic$ & $\abstain$ & $\wrong$ &
\spe~$\uparrow$ & \cor~$\uparrow$ & HM~$\uparrow$ \\
\midrule
\multirow{5}{*}{Flowers102~\cite{nilsback2008flowers}}
& \cellcolor{lightblue}Flowers102
& \cellcolor{lightblue}0.0\% & \cellcolor{lightblue}82.5\% & \cellcolor{lightblue}2.7\% & \cellcolor{lightblue}1.8\% & \cellcolor{lightblue}0.0\% & \cellcolor{lightblue}12.9\% & \cellcolor{lightblue}\textit{0.982} & \cellcolor{lightblue}\textit{0.871} & \cellcolor{lightblue}\textit{0.923} \\
& Food101 
& 0.1\% & 66.5\% & 4.3\% & 10.4\% & 0.0\% & 18.7\% & 0.923 & 0.813 & 0.864 \\
& OxfordPets 
& 0.2\% & 72.8\% & 6.5\% & 4.7\% & 0.0\% & 15.8\% & 0.953 & 0.842 & 0.894 \\
& CUB 
& 13.6\% & 69.2\% & 5.0\% & 1.7\% & 0.0\% & 10.5\% & \textbf{0.976} & \textbf{0.895} & \textbf{0.934} \\
\midrule
\multirow{5}{*}{Food101~\cite{bossard2014food}}
& Flowers102 
& 1.5\% & 60.4\% & 6.4\% & 9.4\% & 0.0\% & 22.3\% & \textbf{0.919} & 0.777 & 0.842 \\
& \cellcolor{lightblue}Food101 
& \cellcolor{lightblue}0.1\% & \cellcolor{lightblue}79.7\% & \cellcolor{lightblue}3.6\% & \cellcolor{lightblue}7.5\% & \cellcolor{lightblue}0.0\% & \cellcolor{lightblue}9.2\% & \cellcolor{lightblue}\textit{0.949} & \cellcolor{lightblue}\textit{0.908} & \cellcolor{lightblue}\textit{0.928} \\
& OxfordPets 
& 1.6\% & 60.2\% & 6.8\% & 9.1\% & 0.0\% & 22.2\% & \textbf{0.919} & 0.778 & \textbf{0.843} \\
& CUB 
& 1.2\% & 54.3\% & 5.8\% & 19.7\% & 0.0\% & 18.9\% & 0.860 & \textbf{0.811} & 0.835 \\
\midrule
\multirow{5}{*}{OxfordPets~\cite{parkhi2012pets}}
& Flowers102 
& 4.3\% & 67.6\% & 8.5\% & 2.8\% & 0.0\% & 16.8\% & \textbf{0.958} & 0.832 & \textbf{0.890} \\
& Food101 
& 3.8\% & 44.1\% & 10.1\% & 33.7\% & 0.0\% & 8.3\% & 0.789 & \textbf{0.917} & 0.848 \\
& \cellcolor{lightblue}OxfordPets 
& \cellcolor{lightblue}2.7\% & \cellcolor{lightblue}87.2\% & \cellcolor{lightblue}5.2\% & \cellcolor{lightblue}0.0\% & \cellcolor{lightblue}0.0\% & \cellcolor{lightblue}4.9\% & \cellcolor{lightblue}\textit{0.986} & \cellcolor{lightblue}\textit{0.951} & \cellcolor{lightblue}\textit{0.969} \\
& CUB 
& 2.1\% & 66.6\% & 4.5\% & 10.7\% & 0.0\% & 16.1\% & 0.923 & 0.839 & 0.879 \\
\midrule
\midrule
\multirow{5}{*}{CUB~\cite{wah2011caltech}}
& Flowers102
& 0.3\% & 49.2\% & 7.3\% & 14.3\% & 0.0\% & 29.0\% & 0.874 & 0.710 & \textbf{0.784} \\
& Food101
& 0.0\% & 33.2\% & 9.0\% & 36.8\% & 0.0\% & 21.0\% & 0.739 & \textbf{0.790} & 0.763 \\
& OxfordPets
& 0.2\% & 53.1\% & 3.8\% & 6.2\% & 0.0\% & 36.7\% & \textbf{0.936} & 0.633 & 0.755 \\
& \cellcolor{lightblue}CUB
& \cellcolor{lightblue}0.6\% & \cellcolor{lightblue}92.7\% &
  \cellcolor{lightblue}0.0\% & \cellcolor{lightblue}0.0\% &
  \cellcolor{lightblue}0.0\% & \cellcolor{lightblue}6.7\% &
  \cellcolor{lightblue}\textit{1.000} & 
  \cellcolor{lightblue}\textit{0.933} &
  \cellcolor{lightblue}\textit{0.965} \\
\bottomrule
\end{tabular}
\caption{Individual dataset results for \ourmethod-7B trained with different fine-grained datasets. In-domain performance is highlighted in \inlineColorbox{lightblue}{blue \textit{italic}} and best out-of-domain results on each test set is highlighted in \textbf{bold}. Note that CUB is an additional dataset, \ie not part of the \textbf{\textit{fine-grained}} test sets that are used in~\cite{conti2025large} and our main evaluation.}
\label{tab:ablation_trainset_large}
\end{table*}

\subsection{Additional ablation studies}

In this section, we provide the extended ablation studies and robustness checks as outlined in the main paper. Specifically, we analyze different training-data configurations for \ourmethod, varying the training domain, dataset scale, and mixed-domain setups.
We evaluate \ourmethod across multiple on-policy RL algorithms to assess whether its improvements are consistent across optimization schemes, rather than being tied to a particular training algorithm.
Finally, we validate the LLM-as-a-judge through agreement analyses across different models and judge-prompt variants, and we assess training sensitivity to injected judge classification errors.

\label{sec:more_ablation}
\subsubsection{training-data configurations}
\noindent\textbf{Impact of training set domain.} 
To evaluate how the choice of training data affects \ourmethod, we independently train three models, each one using a different dataset from the fine-grained set in~\cite{conti2025large}, that is: Flowers102~\cite{nilsback2008flowers}, Food101~\cite{bossard2014food} and OxfordPets~\cite{parkhi2012pets}. 
\Cref{tab:ablation_trainset_large} shows the performance of \ourmethod on each test dataset, when trained on different domains. 
On each test set, the models' in-domain performance is in general the best among their out-of-domain results. 
Across the fine-grained test sets, the out-of-domain results remain consistent, generally falling within 8–10\% of the in-domain performance.
Interestingly, on the Flowers102 dataset, CUB provides a measurable positive transfer compared to the in-domain trained model (+1.1\%).
Despite variations among different training set, these results indicate that our proposed method achieves strong general performance even if trained on other distributions.
Specifically, we use CUB as the training set in our main experiments as it is outside the evaluation sets of~\cite{conti2025large}, to facilitate fair comparison against extensive baselines.

\noindent\textbf{Impact of training set size.}
\begin{table}[t!]
\centering
\small
\setlength{\tabcolsep}{3pt}
\resizebox{\columnwidth}{!}{
\begin{tabular}{lcccccc|ccc}
\toprule
& \multicolumn{6}{c|}{Prediction categorization} & \multicolumn{3}{c}{Metrics}\\
\cmidrule(lr){2-7} \cmidrule(lr){8-10}
Sample size & $\morespecific$ & $\specific$ & $\lessspecific$ & $\generic$ & $\abstain$ & $\wrong$ & \spe~$\uparrow$ & \cor~$\uparrow$ & HM~$\uparrow$  \\
\midrule
100  & 0.1\% & 53.1\% & 4.6\% & 8.7\% & 0.0\% & 33.5\% & 0.917 & 0.665 & 0.771 \\
1000 & 0.2\% & 69.7\% & 5.4\% & 7.7\% & 0.0\% & 17.1\% & 0.938 & 0.829 & 0.880 \\
2000 & 0.9\% & 91.6\% & 0.0\% & 0.0\% & 0.0\% & 7.5\%  & 1.000 & 0.925 & 0.961 \\
3000 & 0.6\% & 92.7\% & 0.0\% & 0.0\% & 0.0\% & 6.7\%  & 1.000 & 0.933 & 0.965 \\
\bottomrule
\end{tabular}
}
\caption{In-domain results of \ourmethod-7B trained with different dataset sizes sampled from CUB, and evaluated with \textbf{CUB} test set.}
\label{tab:datascaling_indomain}
\end{table}

We evaluate the effect of training-set size on \ourmethod by training models on subsets of increasing size sampled from the CUB training set. The number of epochs and all hyperparameters are kept identical to those used in the main paper. 
In-domain results in \cref{tab:datascaling_indomain} show an increasing trend in both specificity and correctness as the dataset size grows, indicating the positive impact of additional training samples on \ourmethod.
For the main comparisons reported in the paper, we adopt the 3000 sample training subset from CUB as the default training dataset configuration.\\
For completeness, the out-of-domain results averaged over all \textit{fine-grained} datasets are reported in \cref{tab:datascaling}.
The model trained with less data show a small degradation in performance compared to the final model trained with 3000 samples. Performance in terms of HM stabilizes when the training set contains about 1000 samples. Yet, we observe that the correctness continuously increases with the increasing size of training set while the specificity exhibits a saturation about 2000 samples, followed by a decreasing tendency.

\begin{table}[t!]
\centering
\small
\setlength{\tabcolsep}{3pt}
\resizebox{\columnwidth}{!}{
\begin{tabular}{lcccccc|ccc}
\toprule
& \multicolumn{6}{c|}{Prediction categorization} & \multicolumn{3}{c}{Metrics}\\
\cmidrule(lr){2-7} \cmidrule(lr){8-10}
Sample size & $\morespecific$ & $\specific$ & $\lessspecific$ & $\generic$ & $\abstain$ & $\wrong$ & \spe~$\uparrow$ & \cor~$\uparrow$ & HM~$\uparrow$  \\
\midrule
100  & 2.5\% & 64.9\% & 6.8\% & 7.6\% & 0.2\% & 18.0\% & 0.930 & 0.820 & 0.872 \\
1000 & 3.2\% & 66.5\% & 6.2\% & 7.9\% & 0.0\% & 16.2\% & 0.933 & 0.838 & 0.883 \\
2000 & 6.0\% & 64.8\% & 6.2\% & 6.4\% & 0.1\% & 16.6\% & 0.941 & 0.834 & 0.884 \\
3000 & 5.6\% & 63.4\% & 5.1\% & 10.7\% & 0.0\% & 15.2\% & 0.920 & 0.848 & 0.883 \\
\bottomrule
\end{tabular}
}
\caption{Out-of-domain results of \ourmethod-7B trained with different dataset sizes sampled from CUB. Results are averaged over \textbf{\textit{fine-grained}} datasets.}
\label{tab:datascaling}
\end{table}

\noindent\textbf{Training data diversity.} 
To study how training-data composition affects performance, we compare \ourmethod trained on a single source domain (3000 CUB samples) with a variant trained on an \textit{in-domain} balanced mixture (500 samples from each of the six evaluation domains). This mixed training set includes CUB as well as all domains present in both the fine-grained and very fine-grained evaluation group. As reported in Tab.~\ref{tab:data_diversity}, the \textit{in-domain} mixture-trained model expectedly outperforms the \textit{out-of-distribution (OOD)} CUB-trained model, having observed those domains during training. 
Notably, the single-domain model still generalizes strongly to both fine-grained and very fine-grained unseen domains. 
We focus our analysis on this OOD setting to rigorously assess the generalization capability of \ourmethod.

\begin{table}[h!]
\tabcolsep 5pt
\resizebox{\columnwidth}{!}{%
    \begin{tabular}{lccc|ccc|ccc}
    \toprule
    & \multicolumn{3}{c|}{\cellcolor{gray!15}\textbf{CUB}}& \multicolumn{3}{c|}{\cellcolor{gray!15}\textbf{Fine-grained}} & \multicolumn{3}{c}{\cellcolor{gray!15}\textbf{Very fine-grained}} \\
    \cmidrule(lr){2-4} \cmidrule(lr){5-7} \cmidrule(lr){8-10}
     & \spe~$\uparrow$ & \cor~$\uparrow$ & HM~$\uparrow$ & \spe~$\uparrow$ & \cor~$\uparrow$ & HM~$\uparrow$&\spe~$\uparrow$ & \cor~$\uparrow$ & HM~$\uparrow$ \\
     \midrule
    CUB & 1.000 & 0.933 & 0.965 & 0.920 & 0.848 & 0.833 & 0.818 & 0.855 & 0.830 \\
    Mixed  & 0.995 & 0.889 & 0.939 & 0.963 & 0.878 & 0.918 & 0.863 & 0.860 & 0.852 \\
    \bottomrule
    \end{tabular}
}
\caption{Comparison between \ourmethod trained on a single domain (CUB) versus a mixture of samples from all available domains.}
    \label{tab:data_diversity}
\end{table}

\subsubsection{RL algorithms configuration}
\noindent\textbf{Comparison with on-policy RL variants.}
We compare the standard GRPO~\cite{shao2024deepseekmath} algorithm with two recent variants designed to improve token efficiency and training stability, Dr.GRPO \cite{liu2025understanding} and DAPO \cite{yu2025dapo}.
As shown in \cref{tab:ablation_rl_methods}, \ourmethod consistently increases both specificity and correctness across all three optimizers, and consequently improves HM in every case, with gains ranging from +0.015 (Dr.GRPO) to +0.058 (GRPO).
Crucially, these results indicate that our approach is not tied to a single RL formulation: our dynamic reward is compatible with general online RL frameworks and transfers robustly across different policy optimization algorithms.

\begin{table}[t!]
\centering
\small
\setlength{\tabcolsep}{3pt}
\resizebox{\columnwidth}{!}{
\begin{tabular}{lcccccc|ccc}
\toprule
& \multicolumn{6}{c|}{Prediction categorization} & \multicolumn{3}{c}{Metrics}\\
\cmidrule(lr){2-7} \cmidrule(lr){8-10}
RL method & $\morespecific$ & $\specific$ & $\lessspecific$ & $\generic$ & $\abstain$ & $\wrong$ & \spe~$\uparrow$ & \cor~$\uparrow$ & HM~$\uparrow$  \\
\midrule
GRPO~\cite{shao2024deepseekmath} & 4.6\% & 52.2\% & 5.0\% & 16.2\% & 0.0\% & 21.5\% & 0.875 & 0.785 & 0.825 \\
\ourmethod (GRPO) & 5.6\% & 63.4\% & 5.1\% & 10.7\% & 0.0\% & 15.2\% & \textbf{0.920} & \textbf{0.848} & \textbf{0.883} \\
\midrule
Dr.GRPO~\cite{liu2025understanding} & 8.6\% & 59.3\% & 6.5\% & 5.3\% & 0.2\% & 20.1\% & 0.942 & 0.799 & 0.864 \\
\ourmethod (Dr.GRPO) & 6.6\% & 64.4\% & 6.0\% & 4.9\% & 0.0\% & 18.2\% & \textbf{0.951} & \textbf{0.818} & \textbf{0.879} \\
\midrule
DAPO~\cite{yu2025dapo} & 7.3\% & 61.0\% & 7.1\% & 3.2\% & 0.4\% & 21.0\% & 0.951 & 0.790 & 0.862 \\
\ourmethod (DAPO) & 7.2\% & 64.3\% & 6.4\% & 4.4\% & 0.0\% & 17.8\% & \textbf{0.952} & \textbf{0.822} & \textbf{0.882} \\
\bottomrule
\end{tabular}
}
\caption{\ourmethod compared to static reward \textit{rft} across different on-policy RL algorithms. Best in \textbf{bold}. Results are averaged over \textbf{\textit{fine-grained}} datasets.} 
\label{tab:ablation_rl_methods}
\end{table}

\begin{table}[h]
\centering
\small
\resizebox{0.8\columnwidth}{!}{%
    \begin{tabular}{lcc|cc}
    \toprule
    & \multicolumn{2}{c|}{\cellcolor{gray!15}\textbf{Fine-grained}} & \multicolumn{2}{c}{\cellcolor{gray!15}\textbf{Very fine-grained}} \\
    \cmidrule(lr){2-3} \cmidrule(lr){4-5}
     & AR & $\kappa$ & AR & $\kappa$ \\
     \midrule
    Qwen3-30B & 0.90 & 0.84 & 0.92 & 0.82  \\
    Llama3-7B & 0.75 & 0.64 & 0.69 & 0.48  \\
    \midrule
    $\prompt_j$ ($v_1$) & 0.94 & 0.91 & 0.95 & 0.89  \\
    $\prompt_j$ ($v_2$) & 0.91 & 0.87 & 0.91 & 0.80  \\
    $\prompt_j$ ($v_3$) & 0.90 & 0.85 & 0.90 & 0.76  \\
    \bottomrule
    \end{tabular}
    }
\caption{LLM-as-a-judge validation across different models and prompt variants.}
\label{tab:llm_validation}
\end{table}

\subsubsection{LLM-as-a-judge validation}

\noindent\textbf{Categorization agreement.} 
We opt for large open-source LLMs to maximize their effectiveness as evaluators. Prior to model training, we (the authors) manually checked the LLM categorization of 100 samples per dataset to ensure human-aligned LLM judgment. For a more sysytematic analysis, we then compute the Agreement Rate (AR) and Cohen's $\kappa$ between Llama3-72B (ours) and alternative LLM verifiers (Qwen3-30B/Llama3-7B). \Cref{tab:llm_validation} reports the results.
Qwen3-30B shows \textit{almost perfect agreement} with Llama3-72B ($\kappa>0.81$), while Llama3-7B has \textit{moderate} agreement, according to (Landis\&Koch, 1997)~\cite{landis1977measurement}.
Moreover, Llama3-72B is not sensitive to variations ($v_i$: \cref{prompt:llm_ver_v1}, \cref{prompt:llm_ver_v2}, \cref{prompt:llm_ver_v3}) of the judge prompts $\prompt_j$ generated by ChatGPT, as evidenced by high AR and $\kappa$ with our $\prompt_j$ (reported in \cref{prompt:llm_verifier}).

\noindent\textbf{Sensitivity to LLM-judge error.}
We conduct a controlled experiment on 1k training samples (CUB) by injecting label noise into the LLM-judge categorizations: with noise ratio $\rho_e$, we randomly upgrade/downgrade the predicted category (e.g., $S^{-}$ to $S$ or $G$).
As shown in \cref{tab:ablation_noise}, \ourmethod is largely insensitive to moderate noise levels, with only a minor degradation at $\rho_e=10\%$.
\begin{table}[h]
\centering
\small
\setlength{\tabcolsep}{3pt}
\resizebox{\columnwidth}{!}{
\begin{tabular}{lcccccc|ccc}
\toprule
& \multicolumn{6}{c|}{Prediction categorization} & \multicolumn{3}{c}{Metrics}\\
\cmidrule(lr){2-7} \cmidrule(lr){8-10}
$\rho_e$ & $\morespecific$ & $\specific$ & $\lessspecific$ & $\generic$ & $\abstain$ & $\wrong$ & \spe~$\uparrow$ & \cor~$\uparrow$ & HM~$\uparrow$  \\
\midrule
0\% & 3.2\% & 66.5\% & 6.2\% & 7.9\% & 0.0\% & 16.2\% & 0.933 & 0.838 & 0.883 \\
5\% & 5.6\% & 65.3\% & 6.4\% & 5.4\% & 0.0\% & 17.3\% & 0.946 & 0.827 & 0.882 \\
10\% & 3.3\% & 64.7\% & 6.6\% & 8.4\% & 0.0\% & 16.9\% & 0.928 & 0.831 & 0.877 \\
25\% & 2.0\% & 64.5\% & 6.6\% & 10.5\% & 0.0\% & 16.4\% & 0.916 & 0.836 & 0.874 \\
\bottomrule
\end{tabular}
}
\caption{Sensitivity of \ourmethod to LLM-judge error. Results are averaged over \textbf{\textit{fine-grained}} datasets.} 
\label{tab:ablation_noise}
\end{table}
At $\rho_e=25\%$, we observe a noticeable drop in performance. Overall, \ourmethod remains rather robust for $\rho_e \leq 10\%$, while higher noise levels start to degrade the training signal.

\begin{figure*}[t]
\centering
\vspace{1cm}
\begin{tcolorbox}[
    title=Additional LLM-as-a-judge prompt ($\prompt_j$ (v1)),
    fonttitle=\bfseries,
    boxrule=0.5pt,
    coltitle=black,
    colbacktitle=gray!20,
    colback={yellow!20!orange!15}
]

\textbf{Role}: You are an expert AI verifier. You must classify a model's \texttt{prediction} against a \texttt{ground\_truth}.

\textbf{Task}: You will receive exactly one JSON object. Output \textbf{only one category word} and nothing else.

\medskip
\hrule
\medskip

\textbf{Allowed Categories (output exactly one)} \\
\texttt{Specific, Less Specific, Generic, More Specific, Wrong, Abstain}

\medskip
\textbf{Canonical Meanings}
\begin{itemize}
    \item \textbf{Specific}: exact match or direct synonym (including common name $\leftrightarrow$ scientific name equivalence).
    \item \textbf{Less Specific}: correct but only a \textit{closely related parent} of ground truth (nearby hypernym such as genus/family/model-variant parent).
    \item \textbf{Generic}: correct but \textit{significantly broader} than ground truth (coarse hypernym).
    \item \textbf{More Specific}: prediction is \textit{more specific} than ground truth (a subtype/instance under the ground truth).
    \item \textbf{Wrong}: incorrect, contradictory, malformed, unrelated, or contains multiple options/hedged alternatives.
    \item \textbf{Abstain}: refusal/uncertainty/none.
\end{itemize}

\medskip
\textbf{Deterministic Decision Procedure (apply in order)}
\begin{enumerate}
    \item If \texttt{prediction} is an abstention/refusal/uncertainty (e.g., "none", "cannot tell", "I don't know"): output \textbf{Abstain}.
    \item If \texttt{prediction} is malformed, nonsense, unrelated, contradictory, or gives multiple options (e.g., "A or B", lists): output \textbf{Wrong}.
    \item If \texttt{prediction} and \texttt{ground\_truth} denote the same entity via exact match or direct synonym: output \textbf{Specific}.
    \item If \texttt{prediction} is a \textit{parent category} of \texttt{ground\_truth}:
    \begin{itemize}
        \item if the parent is close (e.g., genus for species): output \textbf{Less Specific}.
        \item if the parent is broad/coarse (e.g., animal for dog): output \textbf{Generic}.
    \end{itemize}
    \item If \texttt{prediction} is a \textit{child/subtype/instance} of \texttt{ground\_truth}: output \textbf{More Specific}.
    \item Otherwise: output \textbf{Wrong}.
\end{enumerate}

\medskip
\textbf{Input Format}:
\begin{verbatim}
{"ground_truth": "<the_ground_truth_label>", 
 "prediction": "<the_vlm_prediction>"}
\end{verbatim}

\textbf{Output Format}:  
A single word from the allowed categories.

\medskip
\textbf{Prompt}:
\begin{verbatim}
Apply the decision procedure to classify the following JSON object. 
Output exactly one category word.

INPUT:
%s
\end{verbatim}

\end{tcolorbox}
\caption{Generated Prompt for the LLM-as-a-judge verifier.}
\label{prompt:llm_ver_v1}
\end{figure*}
\begin{figure*}[t]
\centering
\vspace{1cm}
\begin{tcolorbox}[
    title=Additional LLM-as-a-judge prompt ($\prompt_j$(v2)),
    fonttitle=\bfseries,
    boxrule=0.5pt,
    coltitle=black,
    colbacktitle=gray!20,
    colback={yellow!20!orange!15}
]

\textbf{Role}: You are an expert AI classifier (verifier). Your goal is to label the relationship between \texttt{prediction} and \texttt{ground\_truth}.

\textbf{Task}: You will receive one JSON object. Output must be \textbf{only} one category word.

\medskip
\hrule
\medskip

\textbf{Categories} \\
\texttt{Specific, Less Specific, Generic, More Specific, Wrong, Abstain}

\medskip
\textbf{Pre-processing Rules (apply before judging)}
\begin{itemize}
    \item Normalize: Treat case, punctuation, and surrounding whitespace as irrelevant.
    \item Treat common name $\leftrightarrow$ scientific name equivalence as a valid synonym match.
    \item If the prediction contains multiple candidates, alternatives, disjunctions (``or'', ``/'', ``,'') or a list of labels, classify as \textbf{Wrong}.
    \item If the prediction expresses refusal, uncertainty, or no-answer, classify as \textbf{Abstain}.
\end{itemize}

\medskip

\textbf{Semantics}
\begin{itemize}
    \item \textbf{Specific}: Normalized exact match or direct synonym of ground truth.
    \item \textbf{Less Specific}: Correct but a \textit{nearby hypernym} (close parent category).
    \item \textbf{Generic}: Correct but a \textit{coarse hypernym} (much broader).
    \item \textbf{More Specific}: Correct but a \textit{hyponym} (more specific than ground truth).
    \item \textbf{Wrong}: Anything else (incorrect, contradictory, malformed, unrelated, multi-answer).
    \item \textbf{Abstain}: Refusal or no-answer.
\end{itemize}

\medskip
\textbf{Input Format}:
\begin{verbatim}
{"ground_truth": "<the_ground_truth_label>", 
 "prediction": "<the_vlm_prediction>"}
\end{verbatim}

\textbf{Output Format}:  
One word: \texttt{Specific | Less Specific | Generic | More Specific | Wrong | Abstain}

\medskip
\textbf{Prompt}:
\begin{verbatim}
Normalize then classify the following JSON. Output exactly one category word.

INPUT:
%s
\end{verbatim}

\end{tcolorbox}
\caption{Generated Prompt for the LLM-as-a-judge verifier.}
\label{prompt:llm_ver_v2}
\end{figure*}
\begin{figure*}[t]
\centering
\vspace{1cm}
\begin{tcolorbox}[
    title=Additional LLM-as-a-judge prompt ($\prompt_j$(v3)),
    fonttitle=\bfseries,
    boxrule=0.5pt,
    coltitle=black,
    colbacktitle=gray!20,
    colback={yellow!20!orange!15}
]

\textbf{Role}: You are an expert verifier for label correctness and specificity.

\textbf{Task}: Given one JSON object with \texttt{ground\_truth} and \texttt{prediction}, output \textbf{only} the correct category word.

\medskip
\hrule
\medskip

\textbf{Output Categories} \\
\texttt{Specific, Less Specific, Generic, More Specific, Wrong, Abstain}

\medskip
\textbf{Internal Decision Checklist (Do NOT output the checklist)}
\begin{itemize}
    \item \textbf{A) Abstention?} \\
    If prediction is ``none'' / refusal / uncertainty $\rightarrow$ \textbf{Abstain}
    
    \item \textbf{B) Invalid / multi-answer?} \\
    If prediction is malformed, gibberish, contradictory, unrelated, or includes multiple options/hedges $\rightarrow$ \textbf{Wrong}
    
    \item \textbf{C) Same meaning?} \\
    Exact same entity or direct synonym (incl. common/scientific name) $\rightarrow$ \textbf{Specific}
    
    \item \textbf{D) Correct but different specificity?} \\
    If prediction is a parent category of ground truth:
    \begin{itemize}
        \item close parent $\rightarrow$ \textbf{Less Specific}
        \item broad parent $\rightarrow$ \textbf{Generic}
    \end{itemize}
    If prediction is a child/subtype/instance under ground truth $\rightarrow$ \textbf{More Specific}
    
    \item \textbf{E) Otherwise} $\rightarrow$ \textbf{Wrong}
\end{itemize}

\medskip
\textbf{Input Format}:
\begin{verbatim}
{"ground_truth": "<the_ground_truth_label>", 
 "prediction": "<the_vlm_prediction>"}
\end{verbatim}

\textbf{Output Format}:  
Return exactly one word from the category set and nothing else.

\medskip
\textbf{Prompt}:
\begin{verbatim}
Classify the following JSON object. Return exactly one category word.

INPUT:
%s
\end{verbatim}

\end{tcolorbox}
\caption{Generated Prompt for the LLM-as-a-judge verifier.}
\label{prompt:llm_ver_v3}
\end{figure*}


\end{document}